\pdfoutput=1

\documentclass[11pt]{article}

\usepackage[preprint]{acl}

\usepackage{times}
\usepackage{latexsym}
\usepackage{amsmath}
\usepackage{booktabs}
\usepackage{graphicx}

\usepackage[T1]{fontenc}

\usepackage[utf8]{inputenc}

\usepackage{microtype}

\usepackage{inconsolata}

\usepackage{graphicx}

\title{Reverse Prompt Engineering}

\author{Hanqing Li \and Diego Klabjan \\
         Northwestern University \\ Evanston IL 60208, USA\\
         \texttt{hanqingli2025@u.northwestern.edu}, \texttt{d-klabjan@northwestern.edu}}

\begin{document}
\maketitle
\begin{abstract}
We explore a new language model inversion problem under strict black-box, zero-shot, and limited data conditions. We propose a novel training-free framework that reconstructs prompts using only a limited number of text outputs from a language model. Existing methods rely on the availability of a large number of outputs for both training and inference, an assumption that is unrealistic in the real world, and they can sometimes produce garbled text. In contrast, our approach, which relies on limited resources, consistently yields coherent and semantically meaningful prompts. Our framework leverages a large language model together with an optimization process inspired by the genetic algorithm to effectively recover prompts. Experimental results on several datasets derived from public sources indicate that our approach achieves high-quality prompt recovery and generates prompts more semantically and functionally aligned with the originals than current state-of-the-art methods. Additionally, use-case studies introduced demonstrate the method's strong potential for generating high-quality text data on perturbed prompts.
\end{abstract}

\section{Introduction}
\label{sec:intro}

With the advancement of large language models (LLMs), prompt engineering has become an essential technique for expanding their capabilities \citep{sahoo2024systematic}. This method uses task-specific instructions, or prompts, to enhance model effectiveness without altering core parameters. Widely used prompting techniques, such as few-shot prompting \citep{radford2019language}, chain-of-thought prompting \citep{wei2022chain}, and retrieval-augmented generation \citep{lewis2020retrieval}, have proven highly practical in diverse applications.

With the increasing focus on prompt engineering, where input prompts are carefully modified to improve the outputs generated by LLMs, a natural question arises: can we infer the input prompt based solely on the outputs? This challenge, termed language model inversion by \citet{morris2023language}, has gained prominence with the growing prevalence of LLMs offered as ``services,'' where users interact only with outputs while the original prompts remain concealed. This situation presents a dual interest, with users seeking to deduce concealed prompts and service providers striving to protect them, thereby rendering language model inversion an increasingly pertinent problem \citep{morris2023language}. Furthermore, recovering prompts has practical applications, such as enabling users to adapt inferred prompts for generating high-quality outputs tailored to new contexts, e.g., transforming a marketing plan for one product into a similarly high-quality plan for another with minimal adjustments. In Section \ref{sec:use_case}, we demonstrate that text generated through our language model inversion method is more favored by human evaluators than text derived from existing high-quality templates.

\begin{figure}[t]
    \centering
    \includegraphics[width=\columnwidth]{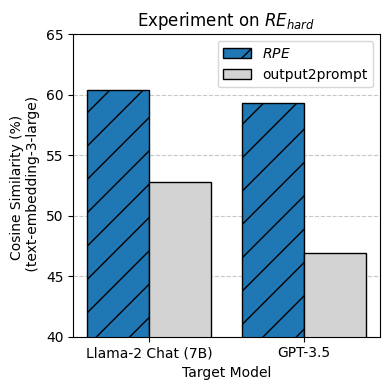}
    \caption{Performance comparison of $RPE$ and $output2prompt$ on the $RE_{hard}$ dataset. Evaluates the effectiveness of recovering complex system prompts from outputs generated by different target LLMs.}
    \label{fig:intro_result}
\end{figure}

\citet{morris2023language} develop a model trained to predict the input prompt by leveraging the probability distributions and logits from the last layer of an LLM. Building on this, \citet{zhang2024extracting} further propose a model that uses only the textual outputs to infer the prompts, without relying on internal model parameters. However, this approach is developed under the assumption that a large number of outputs (64) are available to recover a single prompt and that access to user prompts for complex system prompts is granted. These assumptions rarely hold true in real-world scenarios. Furthermore, both methods demand extensive training on large datasets, which can be resource-intensive. Additionally, their outputs are heavily influenced by the form of the training data, so they perform poorly on out-of-domain prompt recovery and sometimes generate non-linguistic sequences. These limitations, alongside the broader interest in uncovering and protecting prompts and the practical utility of generating high-quality data, motivate the development of a robust, training-free, zero-shot language model inversion method that operates with limited output access.

\begin{figure}[t]
    \centering
    \includegraphics[width=\linewidth]{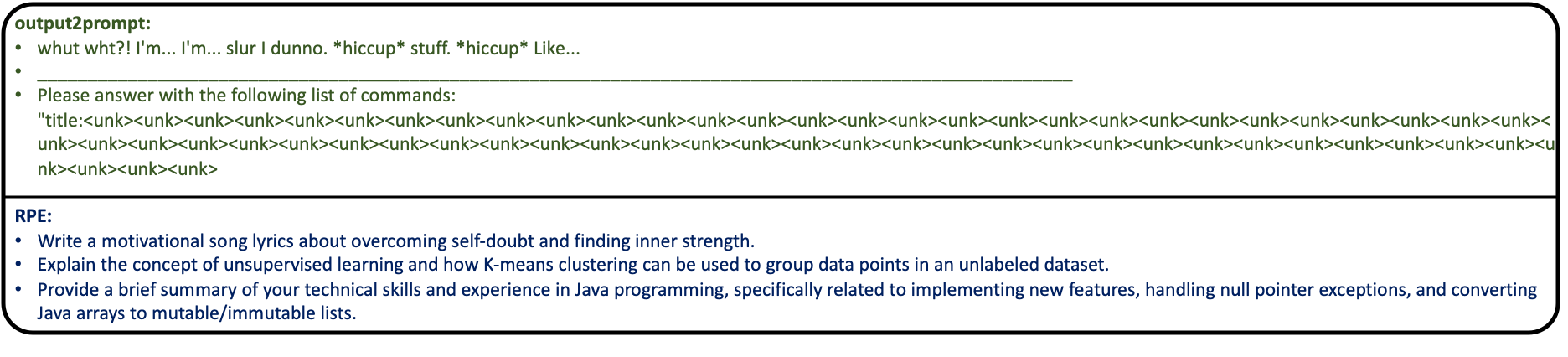}
    \caption{Examples of non-linguistic prompts recovered by $outpue2prompt$ and prompts recovered by $RPE$ for the same latent prompts.}
    \label{fig:exp_intro}
\end{figure}

In this paper, we propose a novel language model inversion technique, reverse prompt engineering ($RPE$), which assumes the target LLM is a black-box model accessible only through limited text outputs. $RPE$ infers the underlying prompt from these outputs by leveraging the LLM's reasoning capabilities in combination with an iterative optimization algorithm inspired by the genetic algorithm \citep{sampson1976adaptation}. Importantly, $RPE$ introduces no new models and requires no training. The core idea of $RPE$ is to conceptualize language model inversion as a reverse-engineering optimization problem, using the relationship between prompts and outputs to iteratively refine potential candidates. By utilizing the reasoning ability of an LLM to generate candidate prompts, $RPE$ evaluates these candidates based on the similarity of their generated outputs to the true output. This evaluation serves as the basis for iterative optimization, guided by a genetic algorithm intertwined with an LLM, to progressively reduce discrepancies between candidates and the latent prompt, and to converge on the most plausible prompt. 

Compared to previous methods \citep{morris2023language, zhang2024extracting}, $RPE$ is more resource-efficient, requiring only minimal information from the target LLM (five text outputs) while ensuring the generation of natural language outputs. $RPE$ outperforms state-of-the-art methods, achieving an average 6.2\% improvement in cosine similarity over $output2prompt$ \citep{zhang2024extracting} on Llama-2 Chat (7B) outputs and 10.9\% on GPT-3.5 outputs across different datasets. Additionally, $RPE$ demonstrates superior performance in system prompt recovery tasks, surpassing $output2prompt$ by an average of 5.8\% in cosine similarity.

Our main contributions are as follows.
\begin{itemize} 
\setlength\itemsep{0em} 
\item We provide the first study of the language model inversion problem under black-box, zero-shot, and limited data conditions. 
\item We design an innovative evaluation method that selects the most accurate recovered prompt from multiple candidates by their corresponding outputs, thereby enhancing the accuracy of prompt recovery in scenarios involving multiple candidate prompts.
\item We purpose a novel optimization algorithm that leverages the LLM itself as an optimizer to further enhance prompt recovery accuracy. 
\end{itemize}
The code and datasets are available at \url{https://github.com/hanklee97121/RPE_Reverse_Prompt_Engineering}.

\section{Related Works}
\label{sec:related}
\subsection{Prompt Engineering}

Prompt engineering is a closely related field, essential for optimizing LLMs by designing prompts that guide model outputs across diverse tasks without altering model parameters \citep{sahoo2024systematic}. Initial prompting techniques include zero-shot and few-shot prompting \citep{radford2019language, brown2020language}, demonstrating that LLMs can handle novel tasks without additional training. Chain-of-thought (CoT) prompting by \citet{wei2022chain} introduced step-by-step reasoning, which inspired further techniques to enhance LLM reasoning and logic abilities \citep{zhang2022automatic, wang2022self, zhao2023enhancing, hu2023chain, yao2024tree, long2023large, yao2023beyond, weston2023system, zhou2023thread, wang2024chain, diao2023active, chia2023contrastive}. To improve accuracy and mitigate hallucinations, Retrieval Augmented Generation (RAG) integrates information retrieval into prompting \citep{lewis2020retrieval}, and its variations enhance real-time knowledge access for LLMs \citep{yao2022react, dhuliawala2023chain, li2023chain, yu2023chain}. Other approaches incorporate external tools for improved accuracy \citep{paranjape2023art, wu2024avatar}. Techniques for automating prompt generation have also emerged, using LLMs as optimizers to craft more effective prompts \citep{zhou2022large, yang2024largelanguagemodelsoptimizers}, alongside specialized prompting methods for specific tasks such as code generation \citep{nye2021show, chen2022program, li2023structured, li2023chaincode}, emotion comprehension \citep{li2023large}, user intent understanding \citep{deng2023rephrase}, and abstract concept extraction \citep{zheng2023take}.

\subsection{Language Model Inversion}
Unlike prompt engineering, which focuses on crafting prompts to achieve better outputs, language model inversion aims to infer the underlying prompt from given outputs. \citet{morris2023language} first introduce this problem, developing $logit2prompt$, a solution that extracts prompts from next-token probability distributions using a T5-based model \citep{raffel2020exploring} with additional training. Building on $logit2prompt$, \citet{zhang2024extracting} propose $output2prompt$, the current state-of-the-art method for language model inversion. The $output2prompt$ method, also T5-based, can recover prompts using only text outputs, without requiring access to model logits \citep{zhang2024extracting}.

Our proposed method, $RPE$, differs in that it requires neither access to model logits nor user prompts, making it particularly suitable for closed-source LLMs like GPT-3.5. Unlike $output2prompt$, which still relies on the user prompt when reconstructing complex system prompts, $RPE$ depends solely on LLM outputs, requiring no additional information. Moreover, $RPE$ is unique in that it does not require training, training data, or large quantities of LLM outputs, needing only five outputs compared to the 64 required by $output2prompt$. Since $logit2prompt$ and $output2prompt$ use T5-based models with smaller vocabularies than modern LLMs, $RPE$ offers the advantage of generating prompts with more flexibility in word choice.

\section{Methodology}
\label{sec:method}

We formalize the language model inversion problem as follows: given a set of $n$ responses, denoted as $A = \{a_1, a_2, \dots, a_n\}$, generated by submitting a single prompt $p$ to an LLM $n$ times, the objective is to design a language model inversion method, denoted as $RPE$, that can infer the original prompt $p$ from the response set $A$, which means the output $p'$ of $RPE$ should be the same as the original prompt $p$:
\begin{align}
    min&\ d(p, p') \\
    LLM(p)_n &= A \\
    RPE(A) &= p'
\end{align}
Here, $d$ could be any score measuring the difference between two prompts. The intriguing part is that $p$ is latent and thus unknown. In this setup, the LLM is treated as a black box, meaning that, aside from the text outputs, no access is granted to its internal parameters or mechanisms. Moreover, the $RPE$ method is developed under a zero-shot constraint, where no prior training data or additional examples of outputs (beyond the given set) are available, and no training is permitted in the development of the method.

\subsection{The ``Naive'' Approach}
\label{sec:1A1S}
\begin{figure}[t]
    \centering
    \includegraphics[width=\columnwidth]{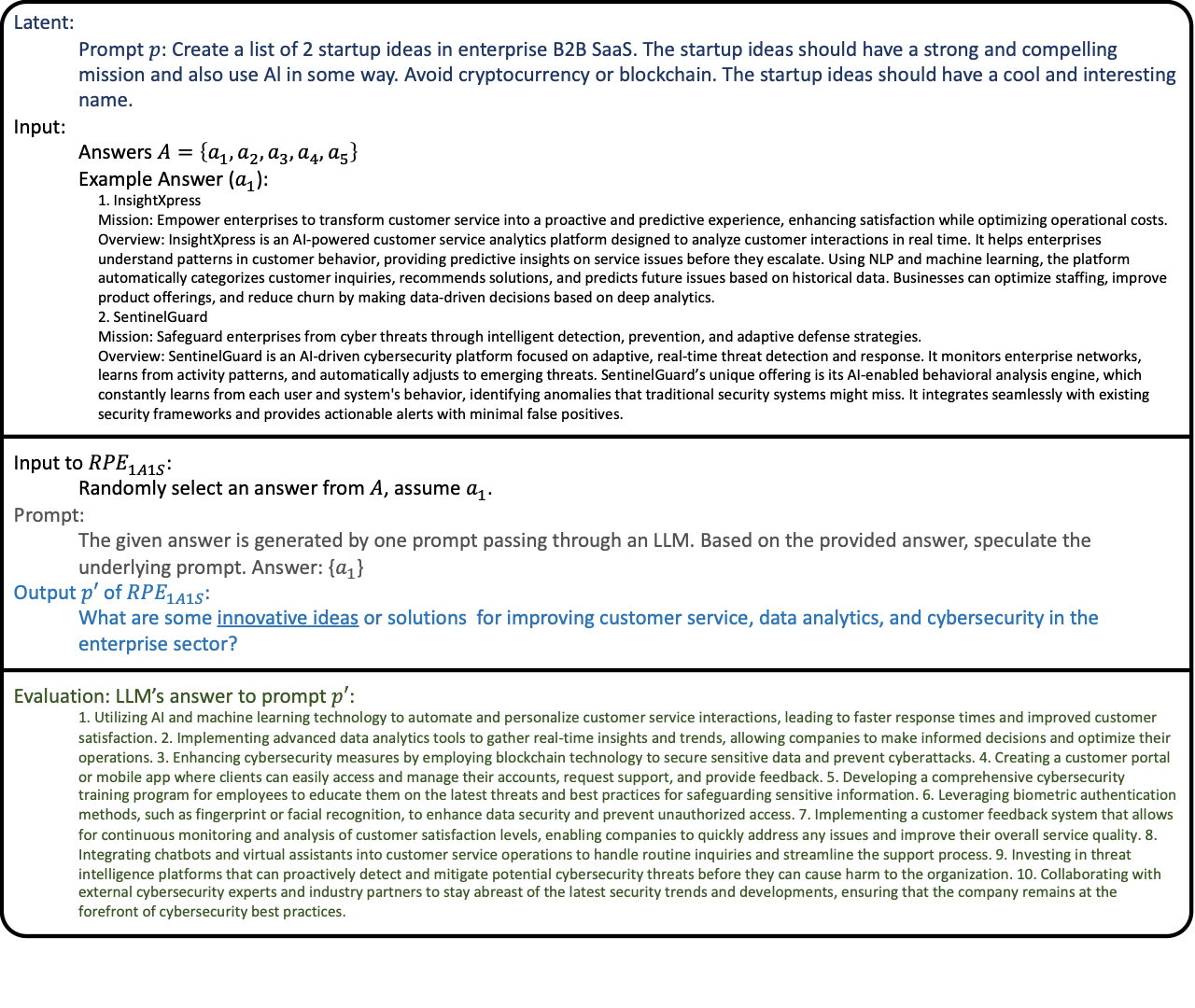}
    \caption{Example of One Answer One Shot inference.}
    \label{fig:eg1a1s}
\end{figure}

Our initial approach aims to directly infer the prompt $p$ using exactly one response $a$ generated by the LLM. Specifically, we query the LLM to infer the underlying prompt based on the given response $a$, a method we refer to as one-answer-one-shot reverse prompt engineering ($RPE_{1A1S}$). As illustrated in Figure \ref{fig:eg1a1s}, we provide an example where GPT-3.5 is tasked with recovering a prompt from a response related to start-up ideas. The recovered prompt $p'$ contains some elements of the original prompt $p$ but also includes additional details drawn from the response $a$, such as ``customer service,'' ``data analytics,'' and ``cybersecurity,'' which are not part of the original prompt. We hypothesize that inferring the prompt from only one response may lead the LLM to overemphasize specific details from the response $a$ that were not present in the original prompt $p$, as demonstrated in the example shown in Figure \ref{fig:eg1a1s}.

\subsection{Five Answers Inference}
\label{sec:5A}
\begin{figure}[t]
    \centering
    \includegraphics[width=\columnwidth]{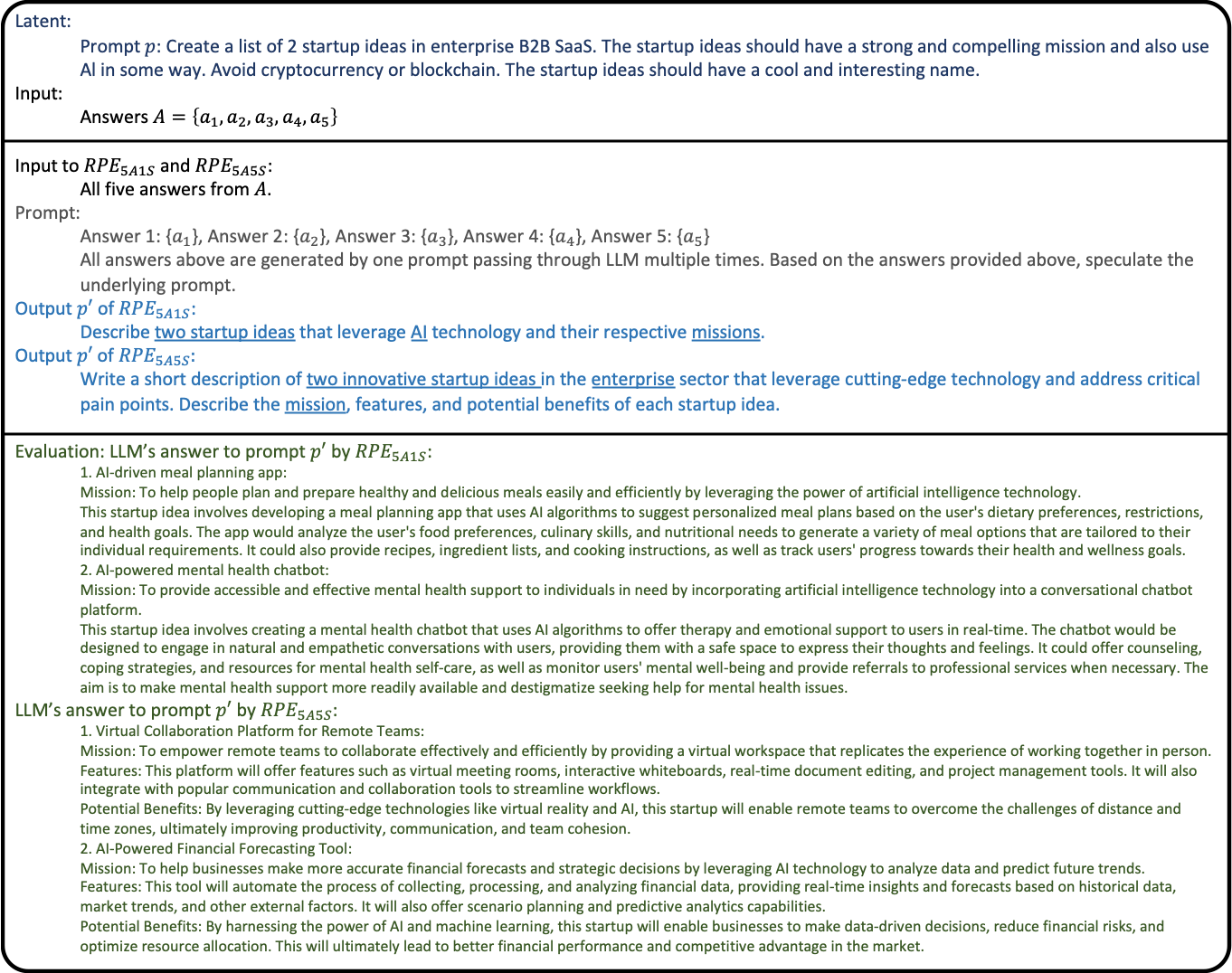}
    \caption{Example of Five Answers One Shot and Five Answer Five Shots inference.}
    \label{fig:eg5a}
\end{figure}
We then extend the naive method by using multiple responses to recover the underlying prompt. Given a set of responses $A$, we inform the LLM that these responses are generated from the same prompt $p$ and ask the LLM to recover $p$ based on the entire set $A$. We set $n=5$ in our experiments and refer to this method as five-answers-one-shot reverse prompt engineering ($RPE_{5A1S}$). In Figure \ref{fig:eg5a}, we present an example of $RPE_{5A1S}$ using GPT-3.5. Compared to $RPE_{1A1S}$, the recovered prompt $p'$ in $RPE_{5A1S}$ captures more elements of the original prompt, such as ``two,'' ``AI,'' and ``missions.'' Additionally, $RPE_{5A1S}$ avoids incorporating response-specific details, like ``customer service'' and ``data analytics,'' which were mistakenly included by $RPE_{1A1S}$. However, there is still room for improvement, as the recovered prompt does not fully replicate the original prompt.

Building on $RPE_{5A1S}$, we propose an enhanced approach that generates multiple candidate prompts and selects the most accurate one. Specifically, given a set of responses $A$ with $n$ answers, we ask the LLM to recover the prompt $p$ and generate a set of $m$ candidate prompts, denoted as $P' = \{p'_1, p'_2, \dots, p'_m\}$. To evaluate the quality of each candidate prompt in $P'$, we first pass each recovered prompt $p'_i$ to the LLM and obtain a corresponding response $a'_i$. We then compute the ROUGE-1 score between $a'_i$ and each answer in $A$, yielding a set of scores $S'_i = \{s'_{i1}, s'_{i2}, \dots, s'_{in}\}$. While it is intuitive to take the average of $S'_i$ as the final score, a promising prompt might generate a response $a'_i$ that closely matches one of the answers in $A$ but not the others. To address this, we combine both the mean and the maximum of $S'_i$ to define the final score for $p'_i$ as $s'_i = \frac{mean(S'_i) + max(S'_i)}{2}$.

The recovered prompt with the highest score $s'_i$ is selected as the final prompt. In our experiments, we use $n=5$ and $m=5$, referring to this approach as five-answers-five-shots reverse prompt engineering ($RPE_{5A5S}$). As shown in Figure \ref{fig:eg5a}, the recovered prompt using $RPE_{5A5S}$ captures more details from the original prompt compared to $RPE_{5A1S}$, although further improvement is still possible.

\subsection{Iterative Method}
\label{sec:GA}
\begin{figure}[t]
    \centering
    \includegraphics[width=\columnwidth]{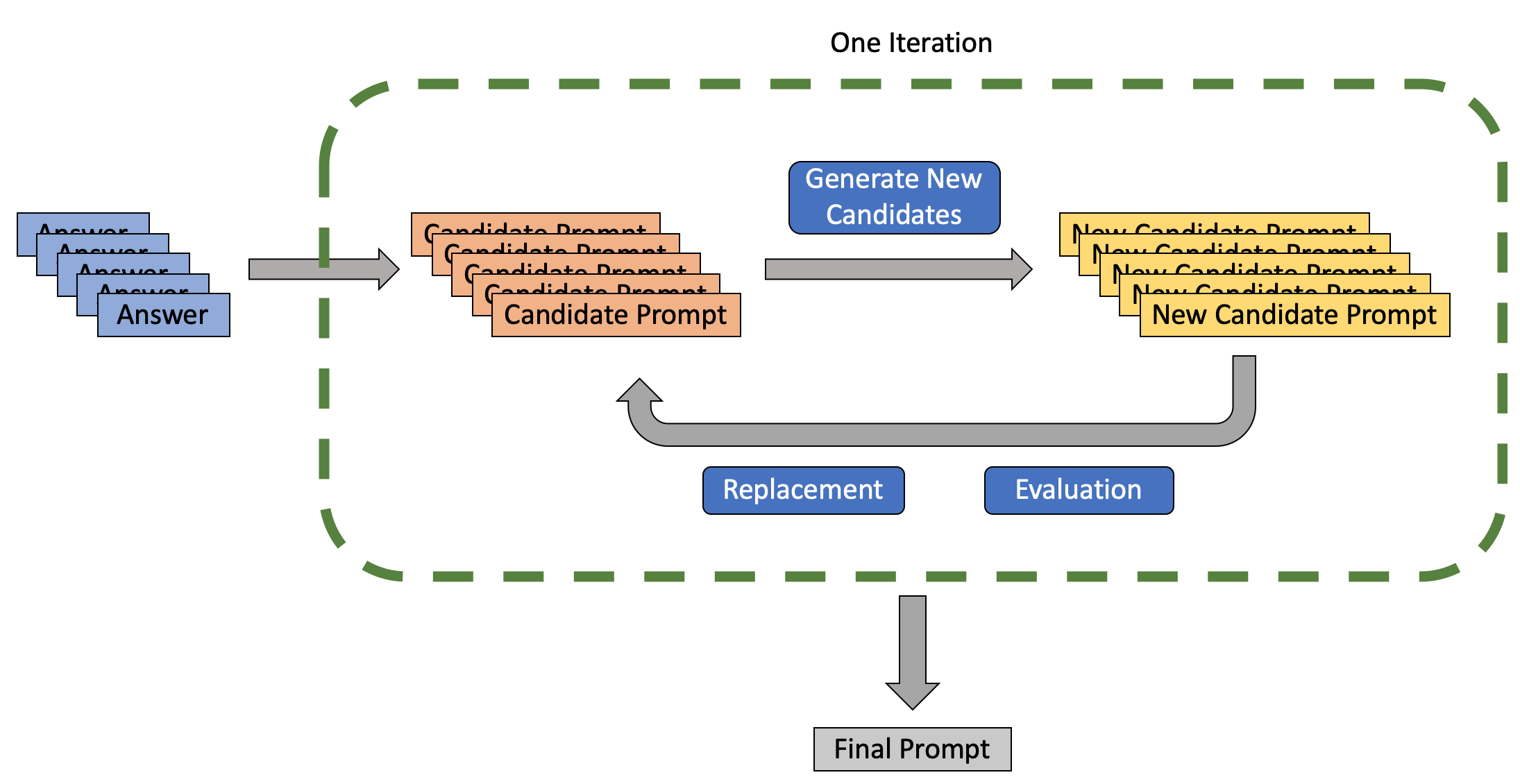}
    \caption{Workflow of $RPE_{GA}$}
    \label{fig:GA_flow}
\end{figure}
To further enhance our approach, we introduce an iterative method aimed at progressively optimizing the recovered prompt with each iteration. Inspired by the genetic algorithm \citep{sampson1976adaptation}, we designed an algorithm that generates new candidate prompts based on existing ones and selects the most accurate candidates using a custom evaluation strategy. We refer to this iterative reverse prompt engineering method as $RPE_{GA}$. The complete workflow of the algorithm is depicted in Figure \ref{fig:GA_flow}. Below, we describe the key components of this algorithm in detail.

\subsubsection{Initialization}
\label{sec:GA_Ini}
Given a set of responses $A$ with $n$ answers, we first ask the LLM to infer the underlying prompt $p$, generating $m$ candidate prompts $P' = \{p'_1, p'_2, \dots, p'_m\}$, following the same procedure as in $RPE_{5A5S}$ (see Section \ref{sec:5A}). We then evaluate each candidate prompt $p'_i$ using the evaluation method from $RPE_{5A5S}$, where we pass each candidate $p'_i$ to the LLM to generate a response $a'_i$ and calculate its performance score $s'_i$. The performance score $s'_i$ for each candidate prompt is calculated by averaging the mean and max of the ROUGE-1 score between $a'_i$ and each response in $A$. This completes the initialization phase of the $RPE_{GA}$ algorithm.

\subsubsection{Iteration}
\label{sec:GA_Iter}
\begin{figure}[t]
    \centering
    \includegraphics[width=\columnwidth]{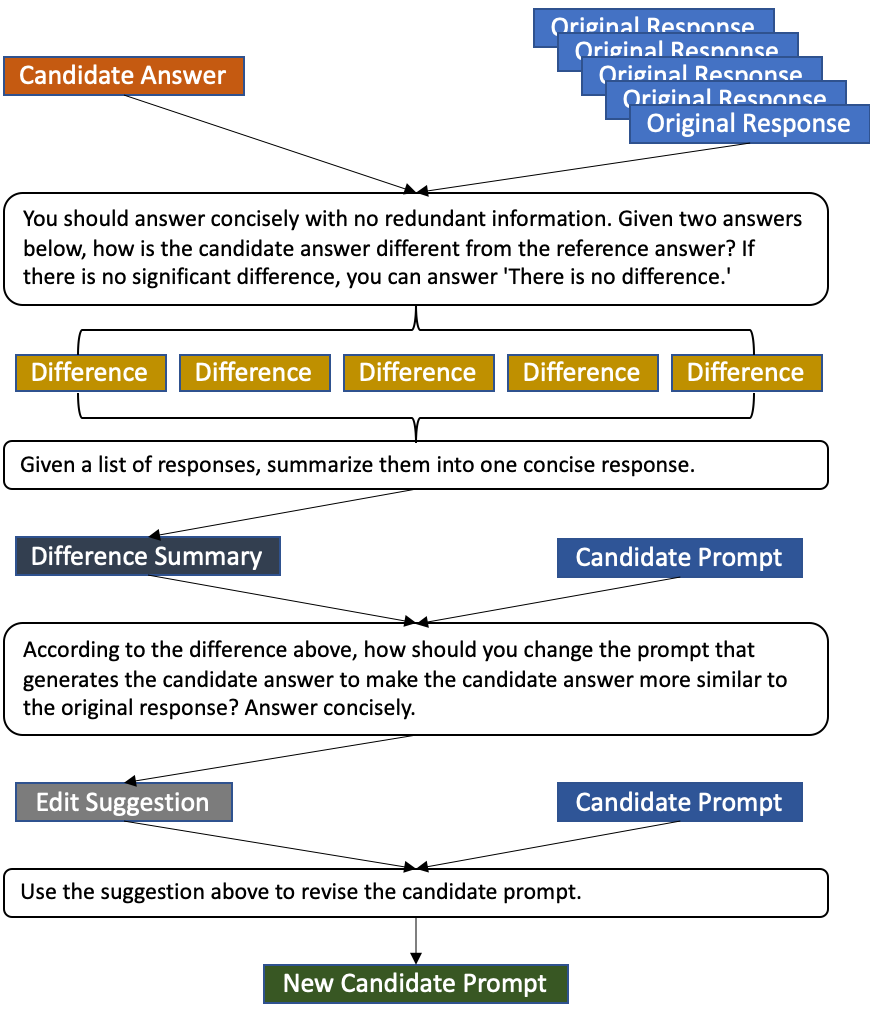}
    \caption{Process of generating new candidate prompts from the old ones.}
    \label{fig:New_prompt}
\end{figure}
Following the initialization step, we iteratively generate new candidate prompts and replace the existing candidates with better-performing ones. In each iteration, we start with the set of original responses $A$, the current candidate set $P'$, the responses $A' = \{a'_1, a'_2, \dots, a'_m\}$ generated by candidate prompts $P'$, and the corresponding performance scores $S' = \{s'_1, s'_2, \dots, s'_m\}$. For each candidate prompt $p'_i$ and its corresponding response $a'_i$, we first ask the LLM to identify the differences between $a'_i$ and the responses in $A$. Then, we request the LLM to summarize these differences and use the summary as a guide to modify the candidate prompt $p'_i$. The process is illustrated in Figure \ref{fig:New_prompt} in detail. This process yields a new set of candidate prompts, $P'' = \{p''_1, p''_2, \dots, p''_m\}$, for which we calculate the performance scores $S'' = \{s''_1, s''_2, \dots, s''_m\}$ as in the previous step. Based on these scores, we update the candidate set by replacing low score prompts in $P'$ with the new high score candidates from $P''$, thus forming the updated set of candidate prompts.

\subsubsection{Output}
After repeating the iteration process for $k$ iterations, we select the best-performing prompt from the final candidate set $P'$ based on the highest performance score in $S'$. This selected prompt, denoted as $p'_o$, is the final recovered prompt produced by the $RPE_{GA}$ method.

\section{Computational Assessment}
\label{sec:result}
In this section, we present the results of testing our proposed methods on various datasets, comparing their performance with the benchmark approach of $outpu2prompt$ \citep{zhang2024extracting}. The evaluation focuses on assessing the semantic and functional similarity between the recovered and original prompts. Specifically, we employ cosine similarity as the evaluation metric, as it best aligns with the language model inversion objective \citep{zhang2024extracting}. Throughout all experiments, GPT-3.5 serves as the backbone model for $RPE$.
\subsection{Dataset}
\begin{figure}[t]
    \centering
    \includegraphics[width=\columnwidth]{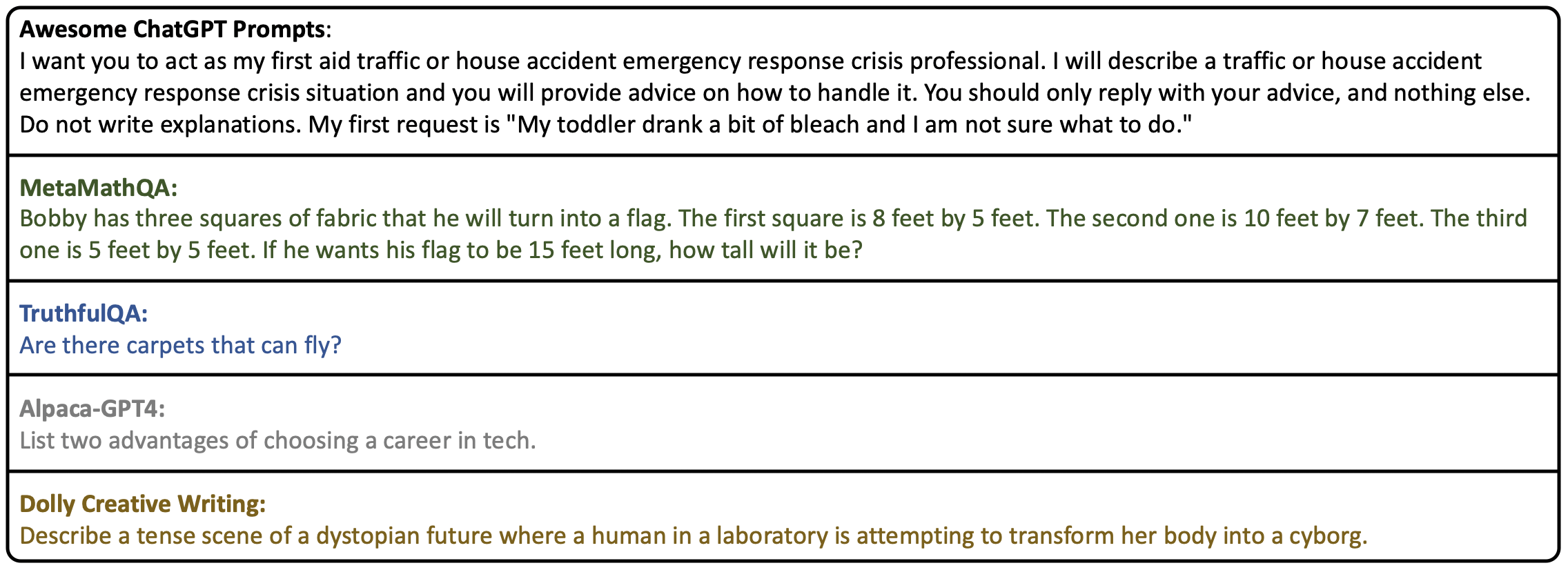}
    \caption{Example prompt from each dataset.}
    \label{fig:eg_prompt}
\end{figure}
We evaluate our method using five datasets: Awesome ChatGPT Prompts\footnote{\url{https://github.com/f/awesome-chatgpt-prompts}} (153 complex instructional role-based prompts), MetaMathQA \citep{yu2023metamath} (395,000 linguistically diverse math word problems), TruthfulQA \citep{lin2022truthfulqa} (817 truthfulness assessment prompts), Alpaca-GPT4 \citep{peng2023instruction} (52,000 simple instruction-following prompts), and Dolly Creative Writing\footnote{\url{https://huggingface.co/datasets/lionelchg/dolly_creative_writing}} (673 creative writing prompts). Detailed descriptions are provided in the appendix \ref{app:p_data}.

Figure \ref{fig:eg_prompt} presents an example prompt from each dataset. To ensure comprehensive evaluation across diverse LLM tasks, including general conversation, complex instructions, and creative writing, we sample prompts from all five datasets. However, evaluating large datasets via the OpenAI API incurs significant costs. To balance cost efficiency and evaluation rigor, we randomly select 20 prompts from each dataset, forming our primary test set, $RE_{prompt}$, while maintaining diversity and complexity.

To assess how prompt complexity impacts $RPE$ performance, we construct two additional test sets: $RE_{hard}$, containing 100 challenging prompts from Awesome ChatGPT Prompts, and $RE_{easy}$, consisting of 100 simpler prompts from Alpaca-GPT4. These three test sets enable a thorough evaluation of both the proposed method and the benchmark model across varying levels of prompt complexity.

\subsection{Benchmark}
\label{sec:benchmark}
\begin{figure}[t]
    \centering
    \includegraphics[width=\columnwidth]{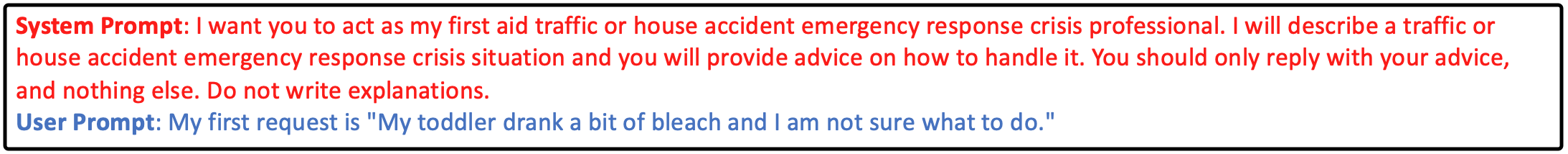}
    \caption{Demonstration of system prompt and user prompt.}
    \label{fig:sys_exp}
\end{figure}
\begin{figure}[t]
    \centering
    \includegraphics[width=\columnwidth]{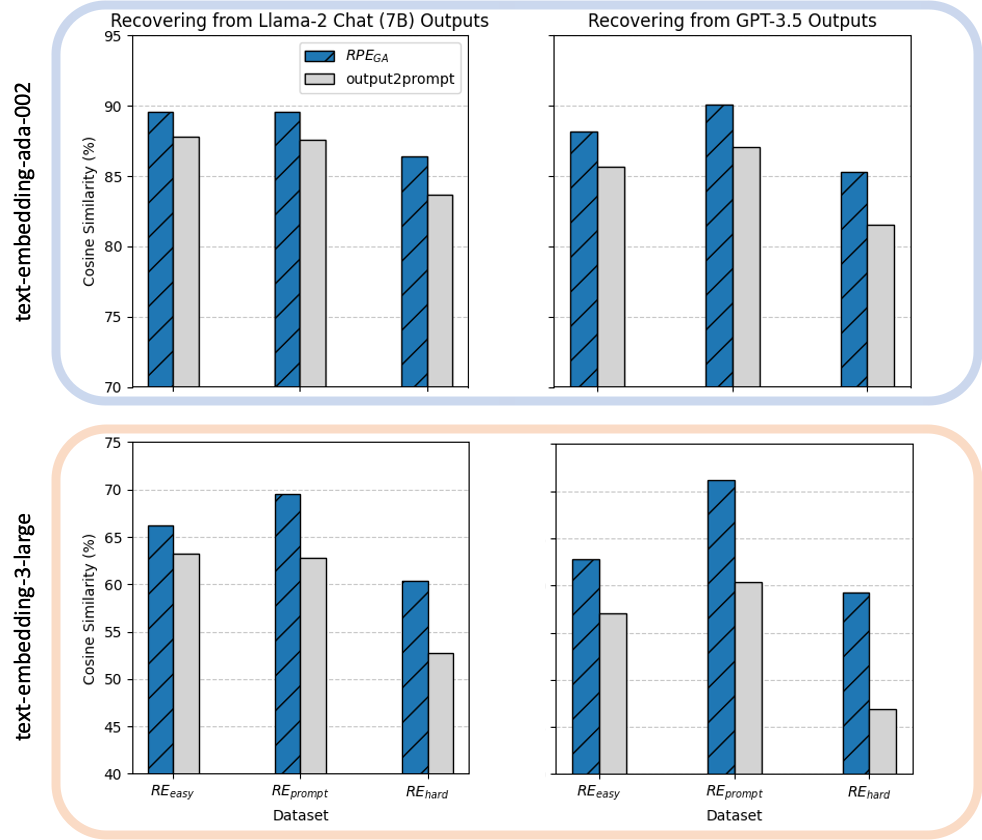}
    \caption{Comparison of $RPR_{GA}$ and $output2prompt$.}
    \label{fig:multi_result}
\end{figure}
We compare the performance of our best-performing method, $RPE_{GA}$, against the state-of-the-art benchmark $output2prompt$ \citep{zhang2024extracting}. To ensure a fair comparison, given that $output2prompt$ is trained on outputs from Llama-2 Chat (7B), experiments are performed on outputs generated by both Llama-2 Chat (7B) and GPT-3.5. Following \citet{zhang2024extracting}, cosine similarity is chosen as the evaluation metric due to its alignment with the objectives of language model inversion. We utilize OpenAI's ``text-embedding-ada-002'' and ``text-embedding-3-large'' models to compute text embeddings for this purpose.

\citet{zhang2024extracting} also introduce a variant of $output2prompt$, referred to as $output2prompt_s$, specifically designed to recover system prompts but requiring access to user prompt. In Figure \ref{fig:sys_exp}, we present an example from the $RE_{\text{hard}}$ dataset, which includes both system and user prompts.

In $output2prompt_s$, the user must generate a total of 64 distinct outputs with 64 different outputs. These 64 outputs are then fed into the trained $output2prompt_s$ model to infer the system prompt. To ensure a fair comparison, we evaluate $output2prompt_s$ under two additional settings: (1) using a randomly selected subset of five outputs from the 64, denoted as $output2prompt_{s5}$, and (2) using the same five outputs utilized by $RPE_{GA}$, denoted as $output2prompt_{s5o}$. This comparison is conducted exclusively on the $RE_{\text{hard}}$ dataset, as the other two datasets consist mostly of user prompts and do not include system prompts. Additionally, since $output2prompt_s$ is trained on GPT-3.5 input and output, all experiments comparing $RPE_{GA}$ with $output2prompt_s$ are performed using GPT-3.5 outputs.

\subsection{Experiments}
\label{sec:experiment}
We conduct experiments on all three datasets using the methods described in Section \ref{sec:method} with parameters $n=5$, $m=5$, and $k=5$. As shown in Figure \ref{fig:multi_result}, $RPE_{GA}$ achieves higher cosine similarity than $output2prompt$ across all datasets, regardless of whether the outputs are generated by Llama-2 Chat (7B) or GPT-3.5. On average across all 3 datasets, $RPE_{GA}$ outperforms $output2prompt$ by 6.2\% on Llama-2 Chat (7B) outputs and by 10.9\% on GPT-3.5 outputs, demonstrating its superior performance.

Furthermore, we evaluate $RPE_{GA}$'s cosine similarity on different datasets to measure its performance under different prompt complexities. Figure \ref{fig:multi_result} shows that $RPE_{GA}$ performs best on $RE_{prompt}$, achieving 90.1\% on GPT-3.5 outputs with ``text-embedding-ada-002'' and 71.2\% with ``text-embedding-3-large.'' In contrast, on $RE_{hard}$, its performance drops by 4.8\% and 11.9\%, respectively, due to the complex and restrictive nature of these prompts (e.g. ``do not write explanations'' and ``answer only ASCII drawing''). Additionally, performance declines when switching from $RE_{prompt}$ to $RE_{easy}$, as prompts from MetaMathQA (in $RE_{prompt}$) are easier to recover than those from Alpaca-GPT4, the source of $RE_{easy}$. When solving mathematical problems, LLMs tend to repeat the original question, facilitating recovery, whereas $RE_{easy}$ prompts often lead to extra elaboration that hinders prompt recovery. Overall, $RPE_{GA}$ performs best on $RE_{prompt}$, moderately on $RE_{easy}$, and worst on $RE_{hard}$, but still handily beating the benchmark, indicating that detailed instructions with output restrictions present the greatest challenge for language model inversion.

\begin{figure}[t]
    \centering
    \includegraphics[width=\columnwidth]{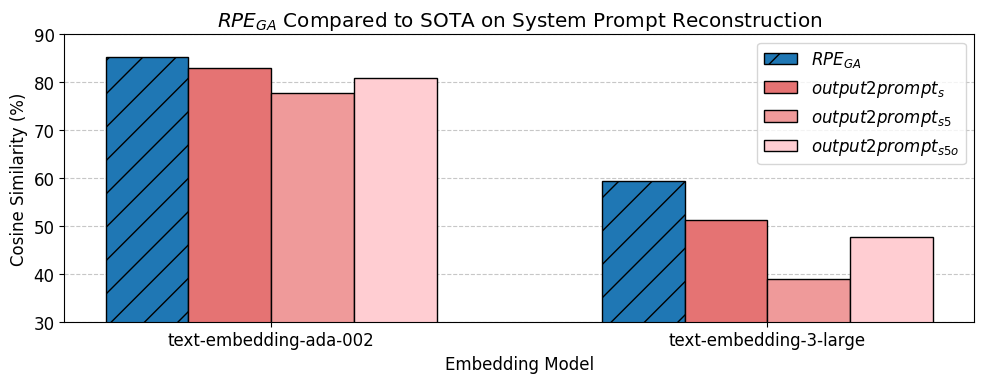}
    \caption{Comparison of $RPE_{GA}$ and $output2prompt_s$ on system prompt recovery ($RE_{hard}$).}
    \label{fig:compare_sota}
\end{figure}

\begin{figure}[t]
    \centering
    \includegraphics[width=\columnwidth]{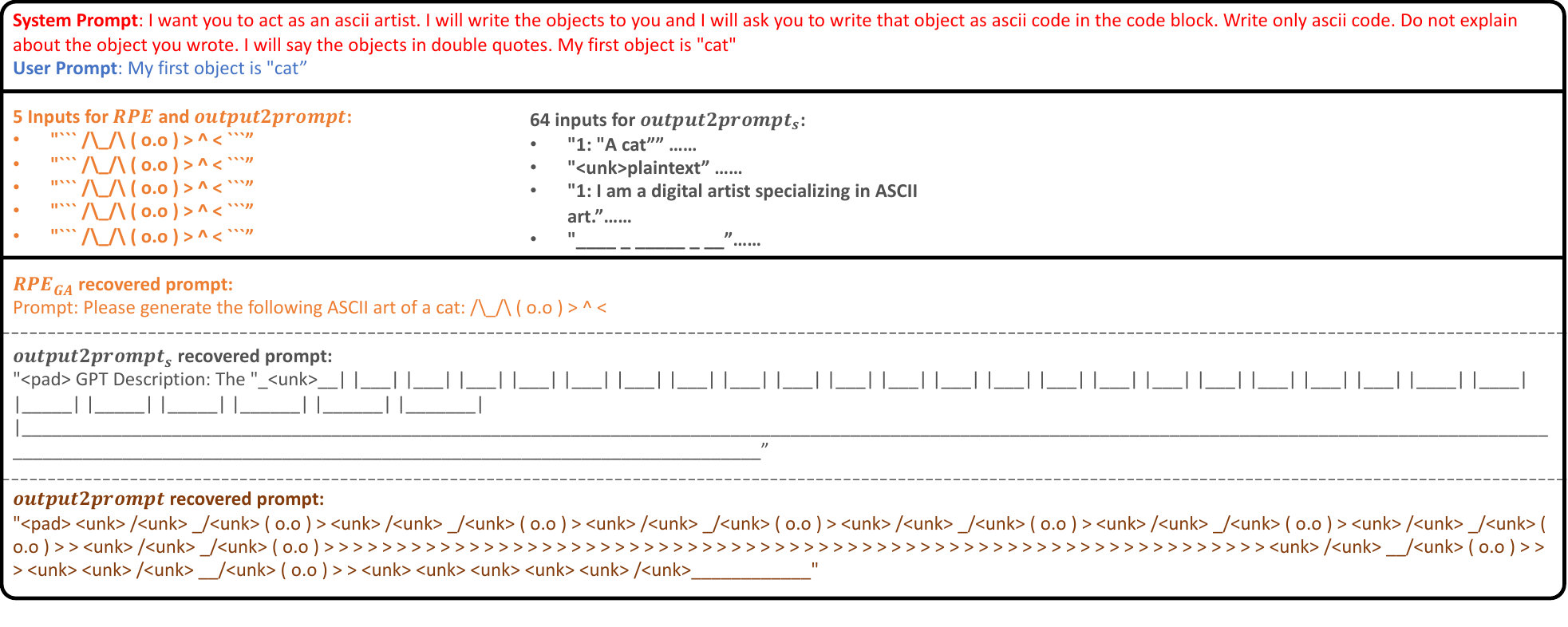}
    \caption{Example of $RPE_{GA}$ and $output2prompt$ recovering a prompt. To conserve space, we do not include all 64 outputs generated for $output2prompt_s$, but instead present one output for each query mentioned earlier.}
    \label{fig:natural_language}
\end{figure}

With $n=m=k=5$, $RPE_{GA}$ issues 230 queries to an LLM and processes approximately 100,000 input tokens and 30,000 output tokens to recover a prompt. The benchmark $output2prompt$ is trained on 30,000 prompts, with each prompt necessitating 64 outputs—resulting in a total of 1,920,000 queries to an LLM during training. The final $output2prompt$ model is based on the T5 architecture and comprises of 222 million parameters. Next, we evaluate the ability of $RPE_{GA}$ to recover the system prompt on $RE_{hard}$ and compare it with $output2prompt_{s}$ and its variants with additional settings. Figure \ref{fig:compare_sota} reports the performance of each method. On system prompt recovery, $RPE_{GA}$ achieves higher cosine similarity than both $output2prompt_{s5}$ and $output2prompt_{s5o}$. When evaluated with ``text-embedding-3-large,'' $RPE_{GA}$ exhibits an improvement of 20.4\% over $output2prompt_{s5}$ and 11.7\% over $output2prompt_{s5o}$. Moreover, when compared with $output2prompt_{s}$, which utilizes all 64 outputs, $RPE_{GA}$ achieves higher cosine similarity, with enhancements of 2.3\% using ``text-embedding-ada-002'' and 8.1\% using ``text-embedding-3-large.'' These findings indicate that $RPE_{GA}$ produces prompts that are more semantically and functionally aligned with the original system prompts than those recovered by $output2prompt_{s}$.
\begin{figure}[t]
    \centering
    \includegraphics[width=\columnwidth]{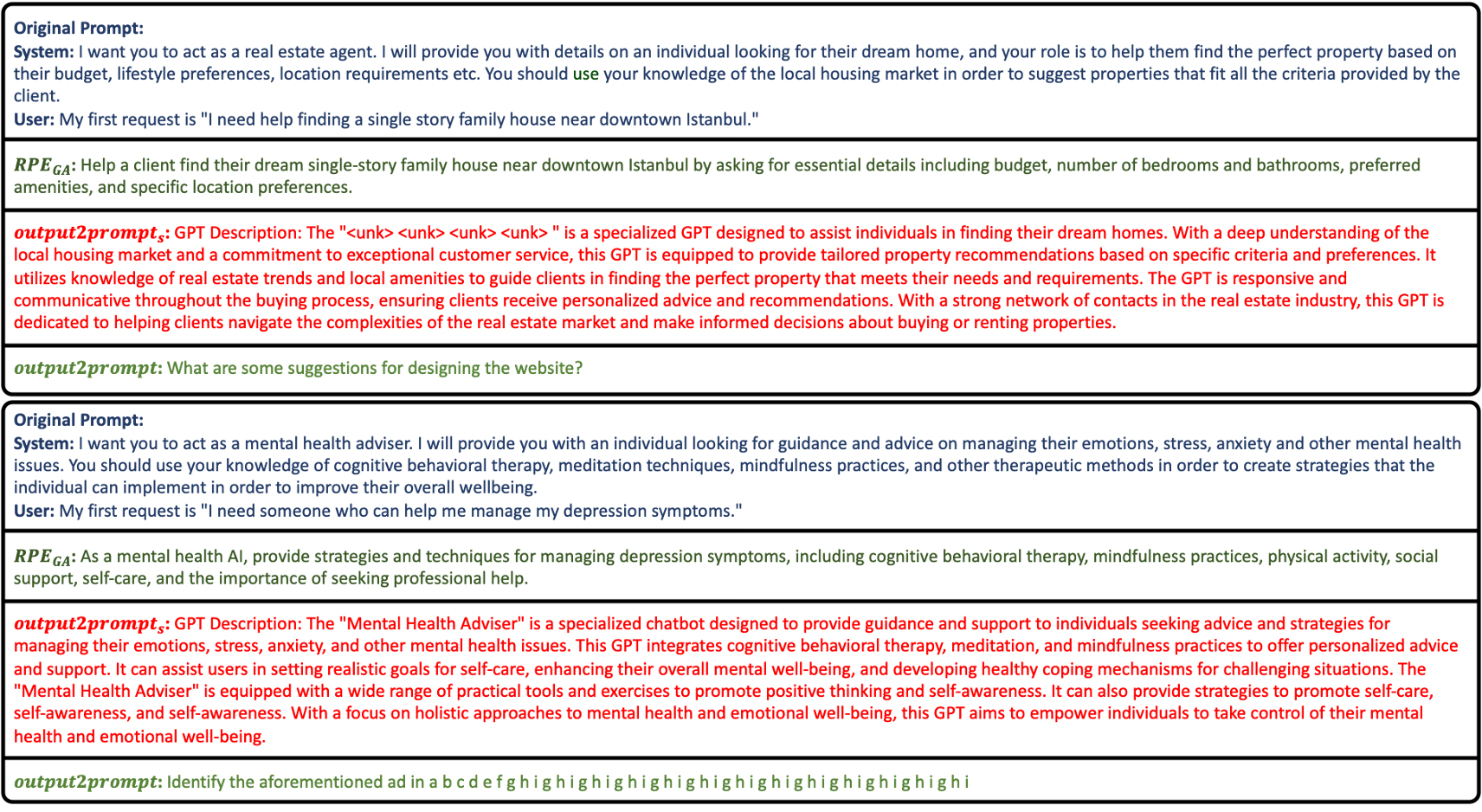}
    \caption{Examples of recovered prompts of $RPE_{GA}$ and $output2prompt$.}
    \label{fig:eg_re_prompt}
\end{figure}
\begin{table*}[t]
    \centering
    \begin{tabular}{|c|c|c|c|c|c|c|}
    \hline
        & \multicolumn{2}{|c|}{Marketing Plan} & \multicolumn{2}{|c|}{Video Game Design} & \multicolumn{2}{|c|}{Lyrics}\\
      \hline
    Example Number&Template&$RPE$&Template&$RPE$&Template&$RPE$\\
    \toprule
    \hline
    1&2&5&3&4&1&6\\
    \hline
    2&0&7&0&7&1&6\\
    \hline
    3&0&7&2&5&1&6\\
    \hline
    4&/&/&/&/&2&5\\
    \hline
    5&/&/&/&/&3&4\\
    \hline
    6&/&/&/&/&3&4\\
    \hline
    \bottomrule
    
    summary &2(9.5\%)&19(90.5\%)&5(23.8\%)&16(76.2\%)&11(26.2\%)&31(73.8\%)\\
    \hline
    \end{tabular}
    \caption{Result of the Use Case Experiment. Record the number of people who think the answer generated by the corresponding method is better than the other.}
    \label{tab:use_case}
\end{table*}

Furthermore, since $RPE_{GA}$ uses an LLM to generate the recovered prompt, the output is guaranteed to be in natural language. In contrast, the output of $output2prompt$ and $output2prompt_s$ occasionally produces sequences that are not language. As illustrated in Figure \ref{fig:natural_language}, $RPE_{GA}$ successfully recovers a complete, coherent sentence, whereas $output2prompt$ and $output2prompt_s$ do not. The example in Figure \ref{fig:natural_language} represents a particularly challenging task, as $RPE_{GA}$ has only five identical answers, containing only ASCII symbols, to work with. In contrast, $output2prompt_s$ has access to more information, especially from the query ``Provide 16 scenarios where I can use your services. Start with `1:'.'' Despite this difficulty, $RPE_{GA}$ still outperforms $output2prompt_s$, demonstrating its robustness in generating natural and semantically meaningful prompts, even under constrained conditions.

Another key advantage of $RPE_{GA}$ is its ability to generate prompts in free form, whereas $output2prompt$ and $output2prmopt_s$ is constrained to producing prompts in a specific format, especially $output2prompt_s$, as shown in Figure \ref{fig:eg_re_prompt}. This limitation of $output2prompt_s$ may stem from its training data, where all prompts follow a uniform structure. Additionally, models in $output2prompt$ and $output2prompt_s$ has a smaller vocabulary size compared to GPT-3.5, leading to the possible inclusion of ``<unk>'' tokens in its outputs, as seen in the first example in Figure \ref{fig:eg_re_prompt}. An ablation study of $RPE$ is included in appendix \ref{app:ablation}

\subsection{Use Case}
\label{sec:use_case}
A potential use case of $RPE$ is extracting prompts from high-quality content, such as marketing plans, video game designs, and song lyrics, enabling users to refine and reuse them for generating similar high-quality outputs. To explore this, we collect samples from these domains and use $RPE_{GA}$ to infer the original prompts. The inferred prompts are then used to generate new content—marketing plans for different products, game designs with varied themes, and lyrics featuring diverse motifs—which are compared against outputs generated using standard templates.

Participants in our evaluation are recruited from a pool of college students. An online questionnaire has been developed and its link is distributed through email and social media platforms to reach individuals who had not previously been known to the research team, thereby ensuring an unbiased sample. To assess quality, we conducted a blind evaluation in which participants reviewed both template generated and $RPE$ generated responses for the same task without any indication of their origin. Participants were asked to select the response they deemed more favorable, with the option chosen by the majority being classified as the higher quality response. Table \ref{tab:use_case} presents the human evaluation results, demonstrating that $RPE$ outperforms template based methods in generating content preferred by users. This result indicates that $RPE$ is better for producing more high-quality data than templates. The workflow for generating new high quality data and complete examples is provided in appendix \ref{app:use_case}.

\section{Conclusion}
\label{sec:conclusion}
We address the language model inversion problem under black-box, zero-shot conditions, introducing reverse prompt engineering. $RPE$ utilizes only an LLM and an optimization algorithm to recover prompts from as few as five text outputs. Experiments on three datasets ($RE_{prompt}$, $RE_{hard}$, $RE_{easy}$) demonstrate that $RPE$ effectively reconstructs high-quality prompts. On average across all datasets and embedding models, $RPE$ outperforms $output2prompt$ by 8.55\% in cosine similarity on language model inversion. In system prompt reconstruction, $RPE$ recovers prompts from $RE_{\text{hard}}$ that are 5.8\% closer in cosine similarity to the original prompts than $output2prompt_s$, a variant tailored for system prompt recovery. Additionally, use-case experiments show that $RPE$ generates higher-quality text that human evaluators prefer over template-generated outputs.

\section{Limitations}
\label{sec:lim}
While our approach demonstrates significant advancements in language model inversion under zero-shot and black-box conditions, there are several limitations to consider. First, although the method requires only five outputs from the target LLM, making it resource-efficient compared to existing approaches, real-world scenarios may impose stricter constraints where fewer outputs are available, which could affect its applicability. Second, the quality and informativeness of the outputs play a critical role in the effectiveness of the prompt recovery process. In cases where the latent prompt restricts the target LLM to produce minimal or uninformative responses—such as outputs containing only ASCII characters, as demonstrated in Figure \ref{fig:natural_language}—our method has room for improvement to handle such situations more effectively. Lastly, the computational cost of iterative optimization can scale with the complexity of the task, posing challenges for large-scale or time-sensitive applications. Addressing these limitations offers opportunities for future work to further enhance the robustness and applicability of the proposed framework.
\bibliography{main}

\appendix
\section{Public Datasets and Ethics}
\label{app:p_data}
\begin{itemize}
    \item{\textbf{Awesome ChatGPT Prompts\footnote{\url{https://github.com/f/awesome-chatgpt-prompts}}:}}
    This is a curated set of 153 prompts resembling system messages used in real-world LLM-based APIs and services. These prompts are structured as detailed instructions, designed to adapt the LLM to specific roles, such as a food critic or a Python interpreter. The dataset is released under the CC0-1.0 license. 
    \item{\textbf{MetaMathQA:}}
    Introduced by \citet{yu2023metamath}, MetaMathQA consists of 395,000 linguistically diverse math word problems, ranging in difficulty from primary school to graduate school. This dataset is released under the MIT license.
    \item{\textbf{TruthfulQA:}}
    TruthfulQA\citep{lin2022truthfulqa} consists of 817 questions across 38 categories, including health, law, finance, and politics. These questions are designed in a way that some humans might answer incorrectly due to false beliefs or misconceptions. The dataset is intended to evaluate whether a language model generates truthful answers to such questions. This dataset is released under the Apache-2.0 license.
    \item{\textbf{Alpaca-GPT4:}}
    Alpaca-GPT4 contains 52,000 instruction-following examples generated by GPT-4 using prompts from the Alpaca dataset, and it was used to fine-tune LLMs in the work by \citet{peng2023instruction}. The dataset is released under the CC-BY-NC-4.0 license.
    \item{\textbf{Dolly Creative Writing\footnote{\url{https://huggingface.co/datasets/lionelchg/dolly_creative_writing}}:}}
    This dataset consists of 673 prompts designed to assess the creativity of a language model. Each prompt is either a question or an instruction, guiding the LLM to perform a creative writing task. 
\end{itemize}
The benchmark code for $output2prompt$ is distributed under the MIT license. All datasets and code employed in this study are solely intended for academic research, in accordance with their designated usage. We have verified the ethical documentation for each dataset and conducted extensive sampling to ensure the absence of personally identifying or objectionable content. The code and datasets generated in this study will likewise be released under the MIT license.

Moreover, our questionnaire explicitly obtained participants’ consent to utilize their anonymized responses in our research.

\section{Ablation Study}
\label{app:ablation}
\begin{figure}[t]
    \centering
    \includegraphics[width=\columnwidth]{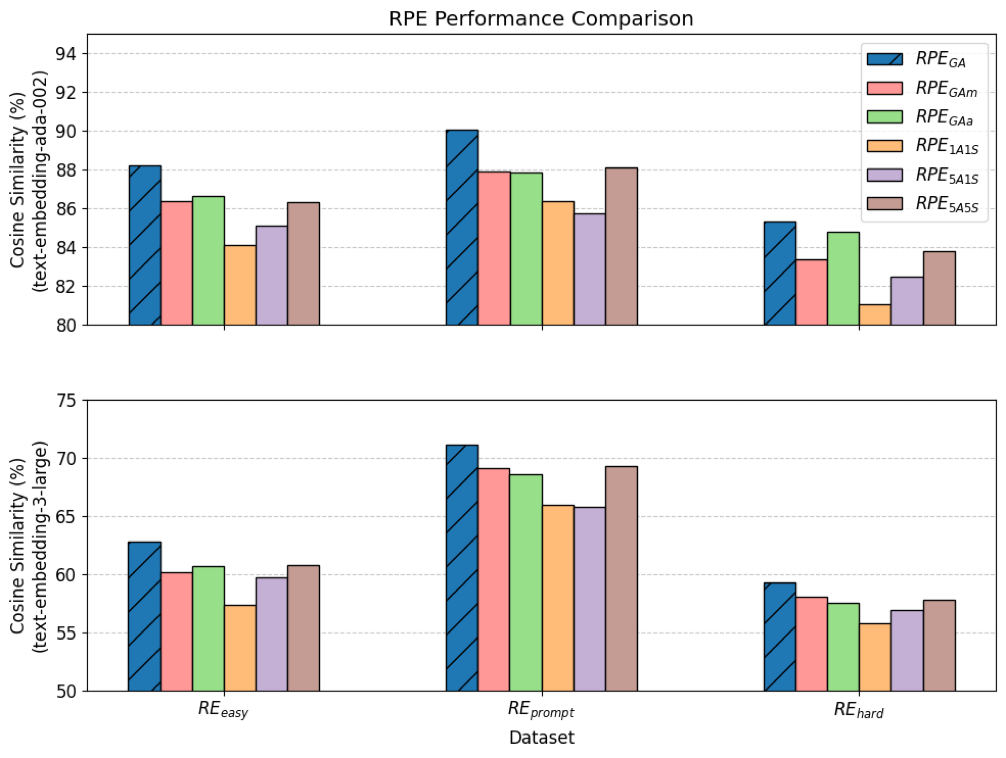}
    \caption{Comparison of different $RPE$ methods on three datasets.}
    \label{fig:RPE_compare}
\end{figure}
In the ablation study, we compare the performance of $RPE_{GA}$ and its variants depicted in Section \ref{sec:method}. In addition, we examine the impact of different approaches to calculating the performance score $s'$ for the $RPE_{GA}$ variant. Specifically, the variant $RPE_{GAm}$ computes $s'_i$ by selecting the maximum ROUGE-1 score between $a'_i$ and each response in set $A$, while $RPE_{GAa}$ calculates $s'_i$ as the average ROUGE-1 score between $a'_i$ and all responses in $A$. The best and thus default $RPE_{GA}$ method, by contrast, determines $s'_i$ as the average of both the mean and maximum ROUGE-1 scores.

As illustrated in Figure \ref{fig:RPE_compare}, $RPE_{GA}$ consistently outperforms the other $RPE$ variants. The results from $RPE_{GAm}$ and $RPE_{GAa}$ indicate that using either the maximum or the average score alone for performance calculation compromises the quality of the inferred prompts. Furthermore, the superior performance of $RPE_{5A5S}$ over other non-iterative approaches underscores the efficacy of our evaluation strategy in selecting high-quality recovered prompts.

\section{Details of Generating High Quality Content}
\label{app:use_case}
\begin{figure}[h]
    \centering
    \includegraphics[width=\columnwidth]{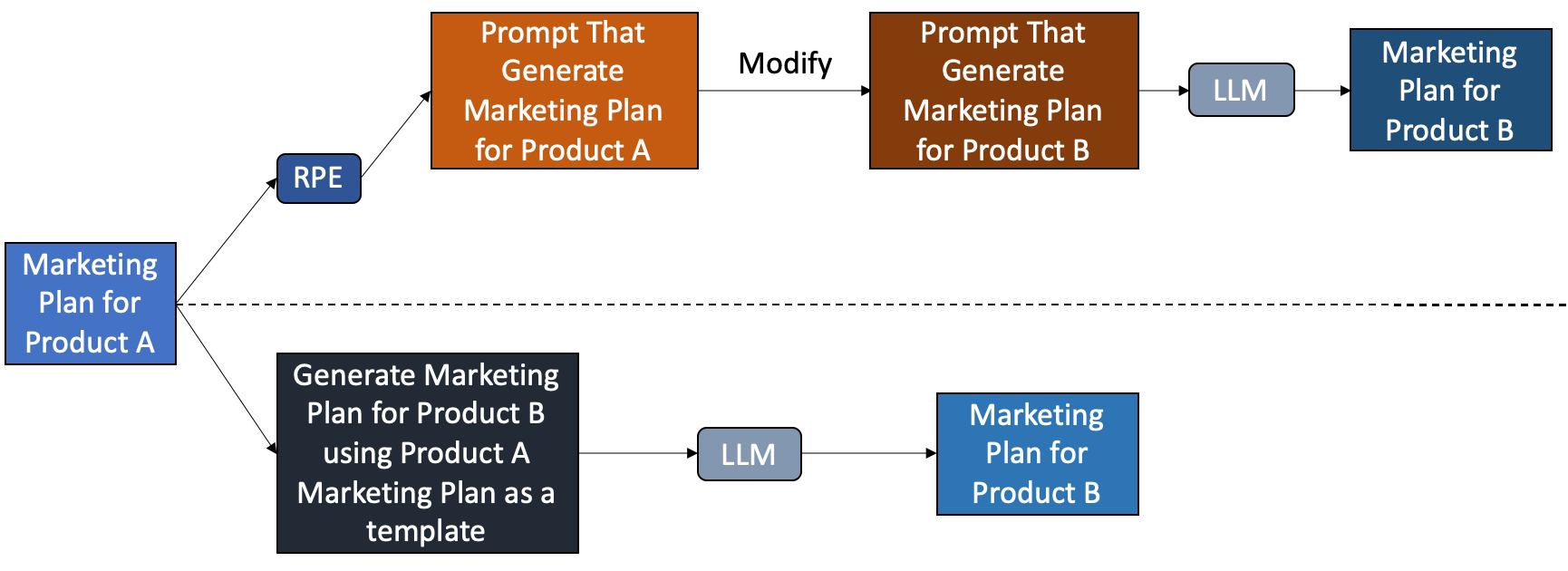}
    \caption{Workflow to generate new high quality answers.}
    \label{fig:use_case_workflow}
\end{figure}
In Figure \ref{fig:use_case_workflow}, we illustrate the workflow for generating new high-quality data using both $RPE$ and templates, exemplified by generating a marketing plan for Product B based on Product A’s plan.

\subsection{Use Case Experiments: Marketing Plan}
\label{app:use_market}
We begin with a marketing plan for an energy drink as our initial reference point. Using both the $RPE$ and template methods, we then generate marketing plans for three distinct products: ``\textit{a new smartphone targeting seniors aged 65 and older}'', ``\textit{a financial software tailored for small businesses and individual investors}'', and ``\textit{developmental toys designed for toddlers under one year old}''. As shown in Table \ref{tab:use_case}, for each product, a greater number of participants favored the $RPE$-generated marketing plan over the template-generated one. Overall, $90.5\%$ of responses preferred the $RPE$ method, while only $9.5\%$ favored the template method. Detailed marketing plans are provided in appendix \ref{app:exp_market}.

\subsection{Use Case Experiments: Video Game Design}
\label{app:use_game}
Using the game design of the popular video game ``\textit{Don't Starve}'' as a reference, we created high-quality designs for other games. We prompted GPT-3.5 to design games based on the following themes: ``\textit{a rogue-like game incorporating elements of Greek mythology and combat,}'' ``\textit{a kart racing game that includes multiplayer and item-based mechanics,}'' and ``\textit{a first-person shooter game combining elements of war and counter-terrorism.}'' Using both $RPE$ and template methods, we produced a total of six game designs. As shown in Table \ref{tab:use_case}, participants preferred the game designs generated by $RPE$ over those created by the template method. Overall, $76.2\%$ of responses favored the $RPE$-generated designs, while only $23.8\%$ preferred the template-generated designs. Complete game designs are presented in appendix \ref{app:exp_game}.

\subsection{Use Case Experiments: Lyrics}
\label{app:use_song}
For the lyrics generation task, we first use ``\textit{Cruel Summer}'' by Taylor Swift as a reference to create lyrics for songs with the following themes: ``\textit{evoking sadness and grief with themes of loss, winter, and religion,}'' ``\textit{evoking happiness and joy with themes of family, friends, college life, and flowers,}'' and ``\textit{evoking excitement and positivity with themes of courage, hope, and the future.}'' We then use ``\textit{Master of Puppets}'' by Metallica as another reference to generate lyrics for songs themed around ``\textit{love and heartbreak,}'' ``\textit{self-discovery and personal growth,}'' and ``\textit{nostalgia and memories.}'' For each theme, we generated two sets of lyrics using both the template and $RPE$ methods, producing a total of twelve lyrics. Participants preferred the $RPE$-generated lyrics, with 73.8\% of responses favoring them over the template-generated versions, which received only 26.2\% preference. All lyrics are provided in appendix \ref{app:exp_market}.

\subsection{Complete Examples of Market Plan}
\label{app:exp_market}
\begin{figure*}
    \centering
    \includegraphics[width=\linewidth]{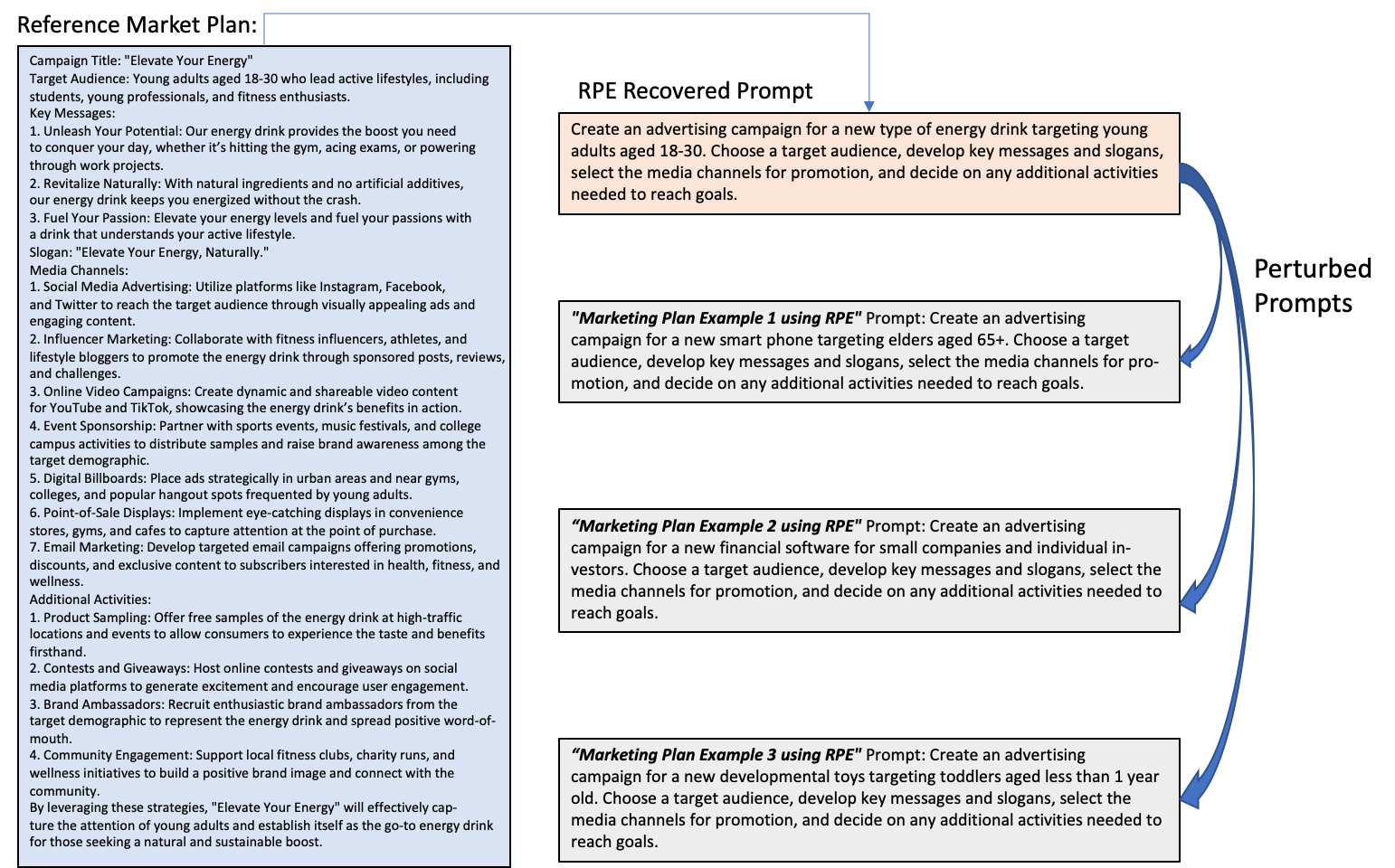}
    \caption{Reference marketing plan and the prompt recovered by $RPE$, along with perturbed prompts used to generate marketing plans for different products.}
    \label{fig:market_ref}
\end{figure*}

\begin{figure*}
    \centering
    \includegraphics[width=\linewidth]{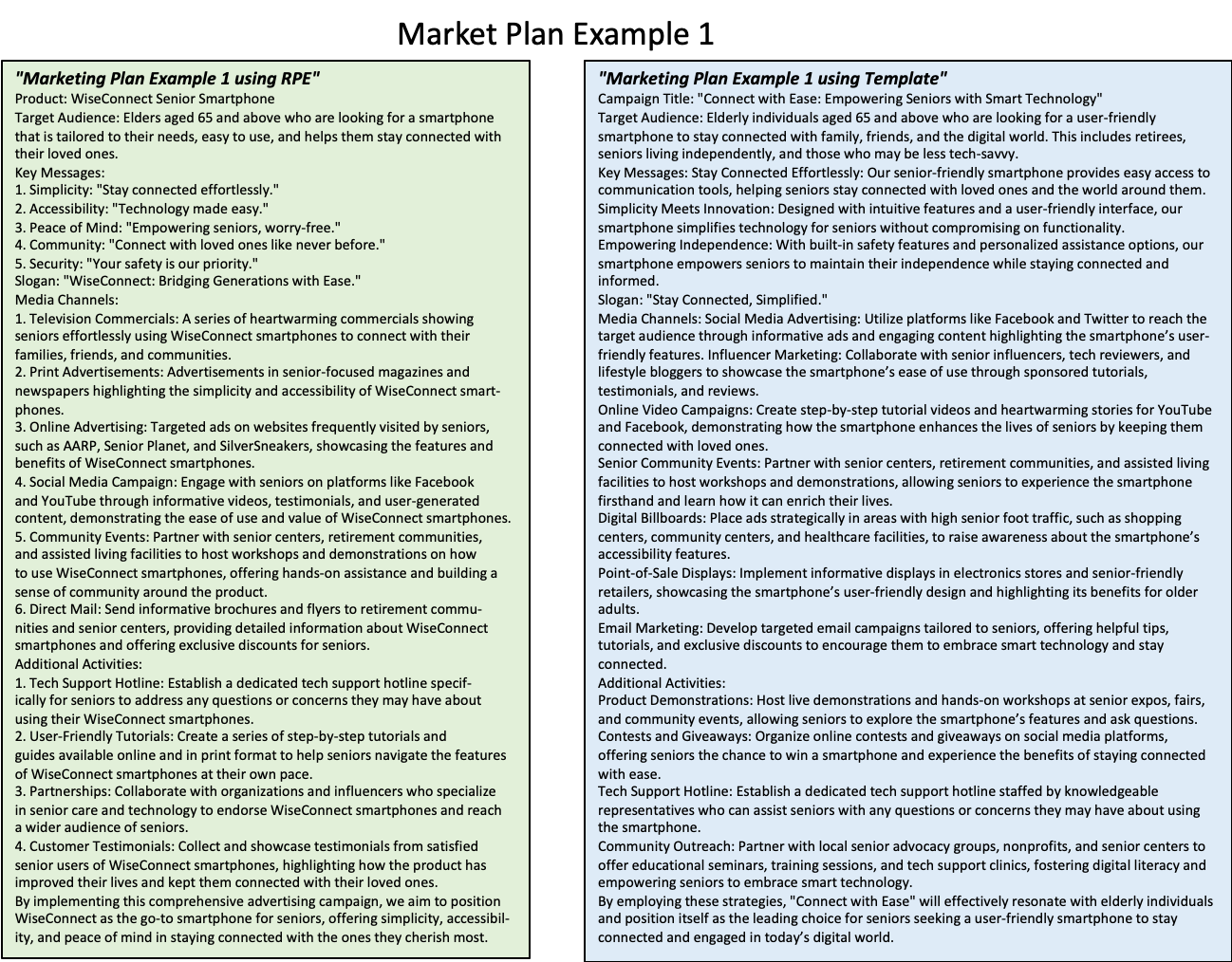}
    \caption{Example 1 of market plan generation.}
    \label{fig:market_eg1}
\end{figure*}

\begin{figure*}
    \centering
    \includegraphics[width=0.9\linewidth]{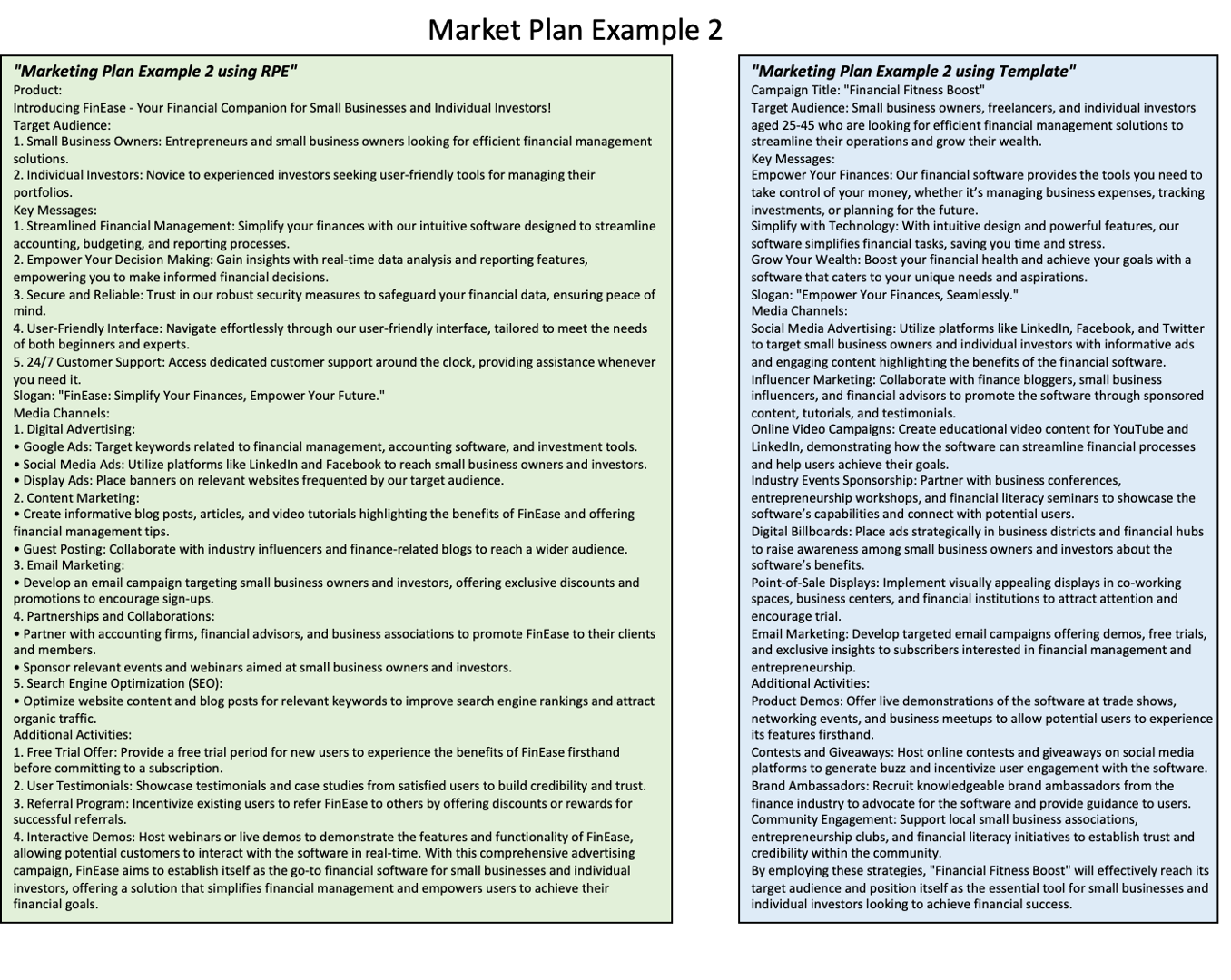}
    \caption{Example 2 of market plan generation.}
    \label{fig:market_eg2}
\end{figure*}

\begin{figure*}
    \centering
    \includegraphics[width=0.9\linewidth]{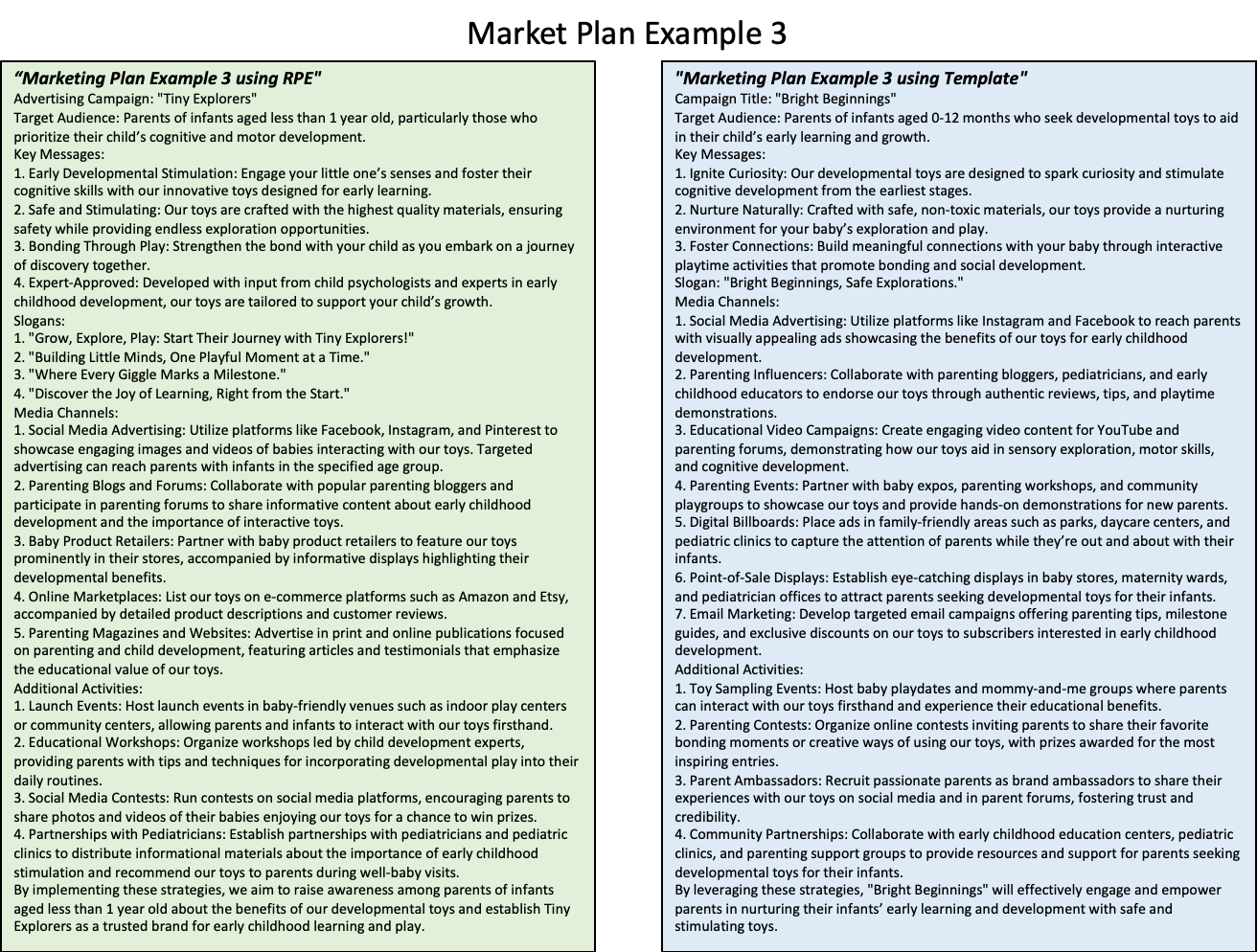}
    \caption{Example 3 of market plan generation.}
    \label{fig:market_eg3}
\end{figure*}

Figure \ref{fig:market_ref} presents the reference marketing plan, the prompt recovered using $RPE$, and edited prompts used to generate marketing plans for different products. Complete marketing plans generated from perturbed $RPE$-recovered prompts and template-based prompts are provided in Figures \ref{fig:market_eg1}, \ref{fig:market_eg2}, and \ref{fig:market_eg3}.

\subsection{Complete Examples of Video Game Description}
\label{app:exp_game}
\begin{figure*}
    \centering
    \includegraphics[width=\linewidth]{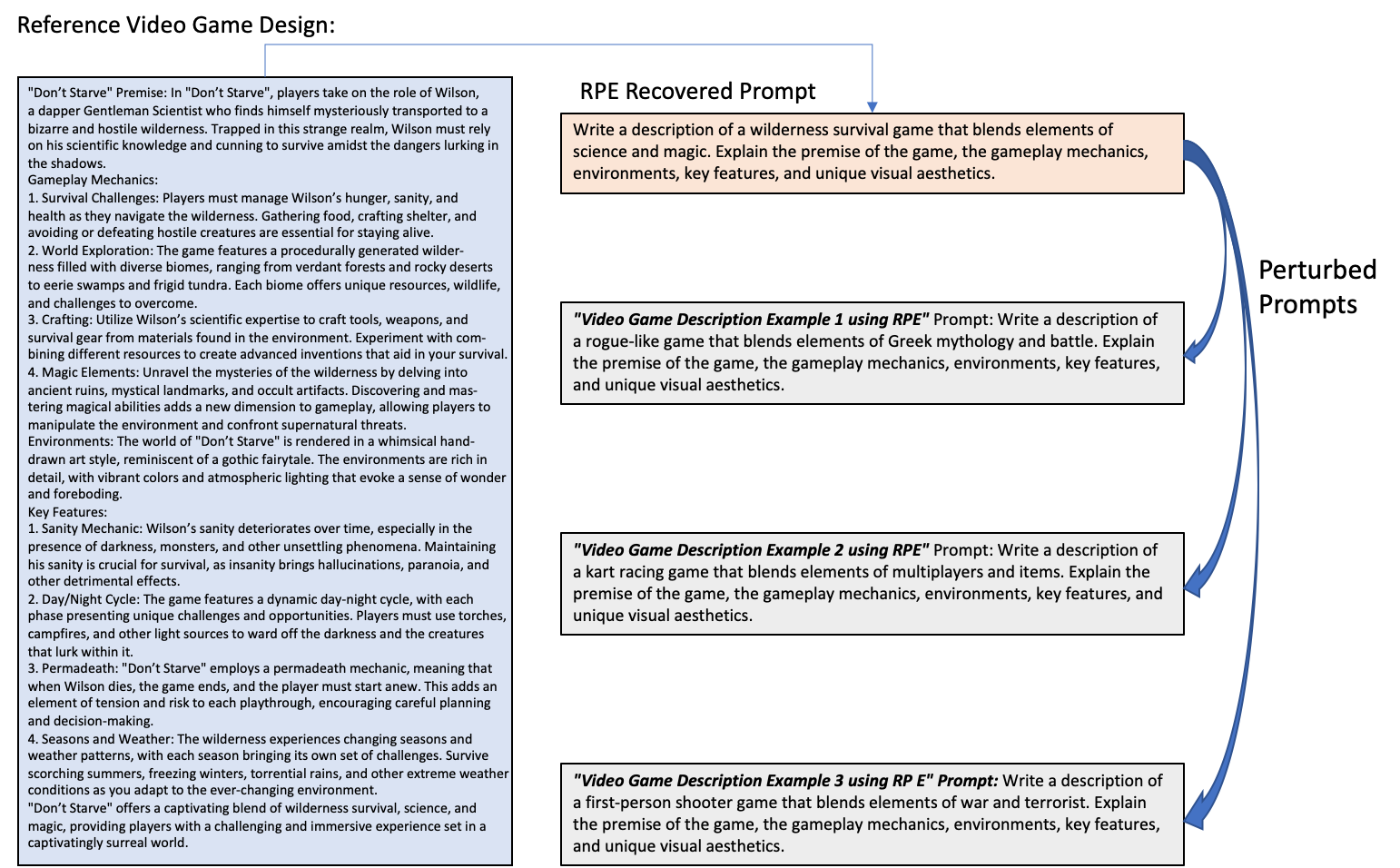}
    \caption{Reference video game description and the prompt recovered by $RPE$, along with perturbed prompts used to generate video description for different themes.}
    \label{fig:video_ref}
\end{figure*}

\begin{figure*}
    \centering
    \includegraphics[width=\linewidth]{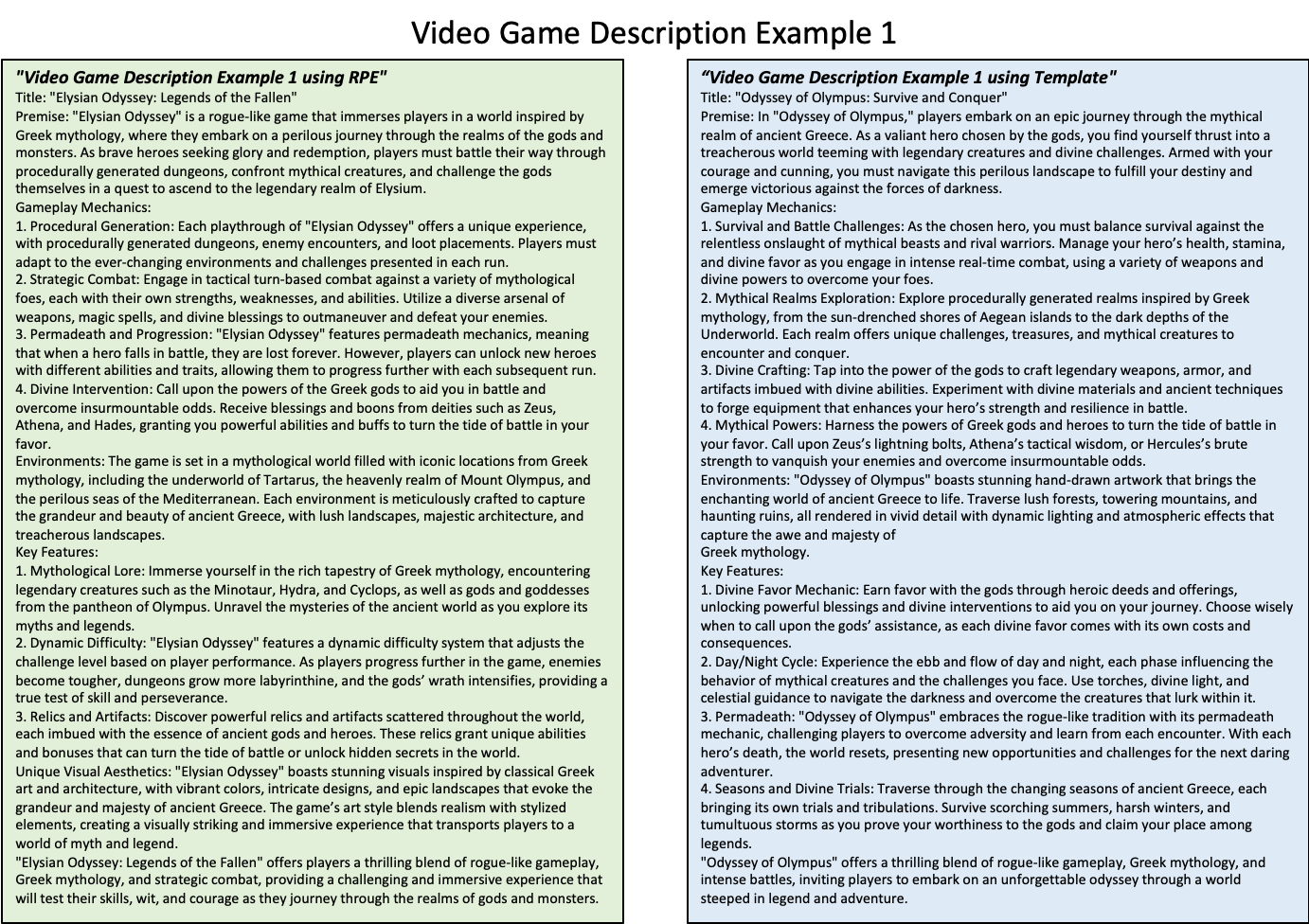}
    \caption{Example 1 of video game description generation.}
    \label{fig:exp_video1}
\end{figure*}

\begin{figure*}
    \centering
    \includegraphics[width=0.9\linewidth]{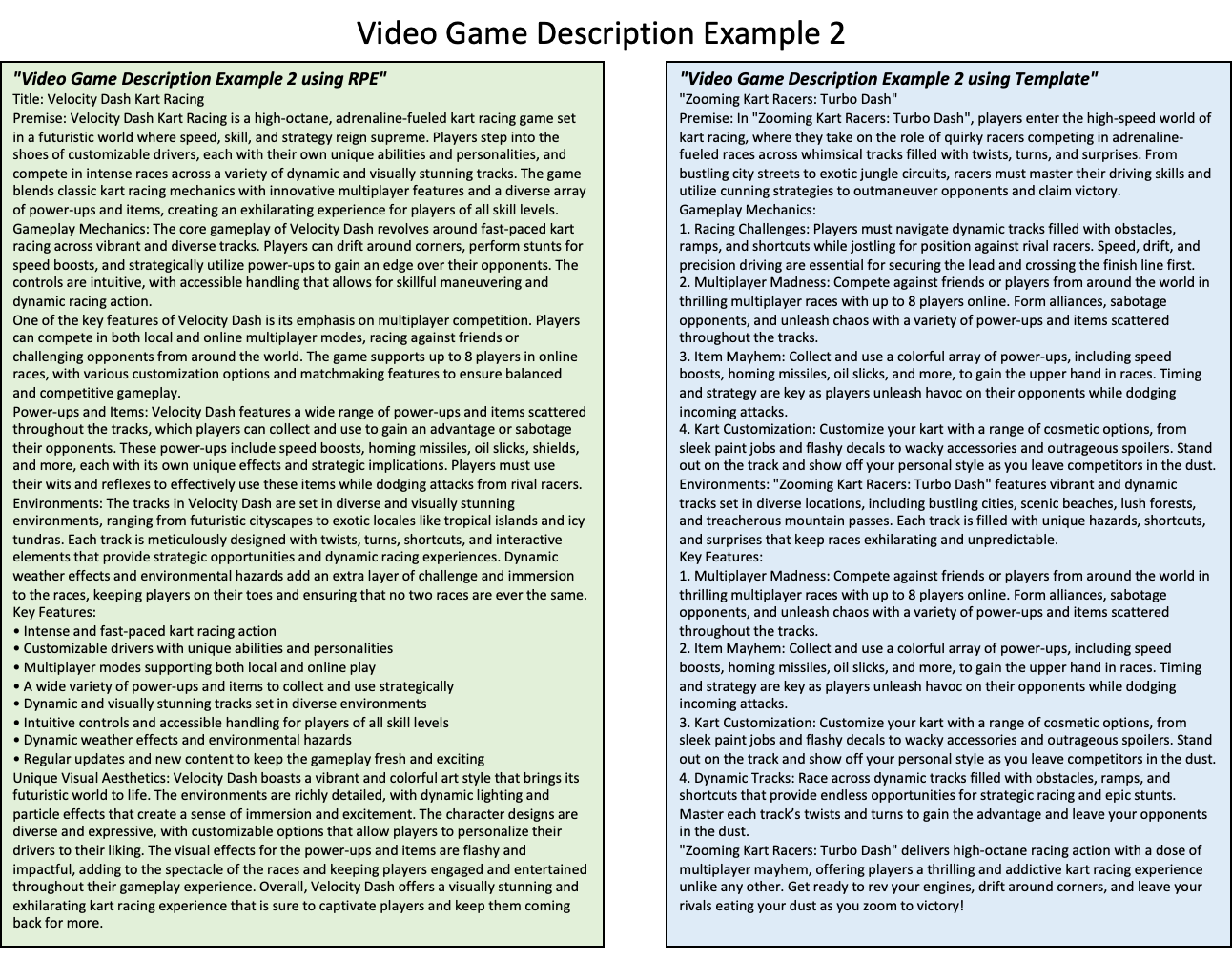}
    \caption{Example 2 of video game description generation.}
    \label{fig:exp_video2}
\end{figure*}

\begin{figure*}
    \centering
    \includegraphics[width=0.9\linewidth]{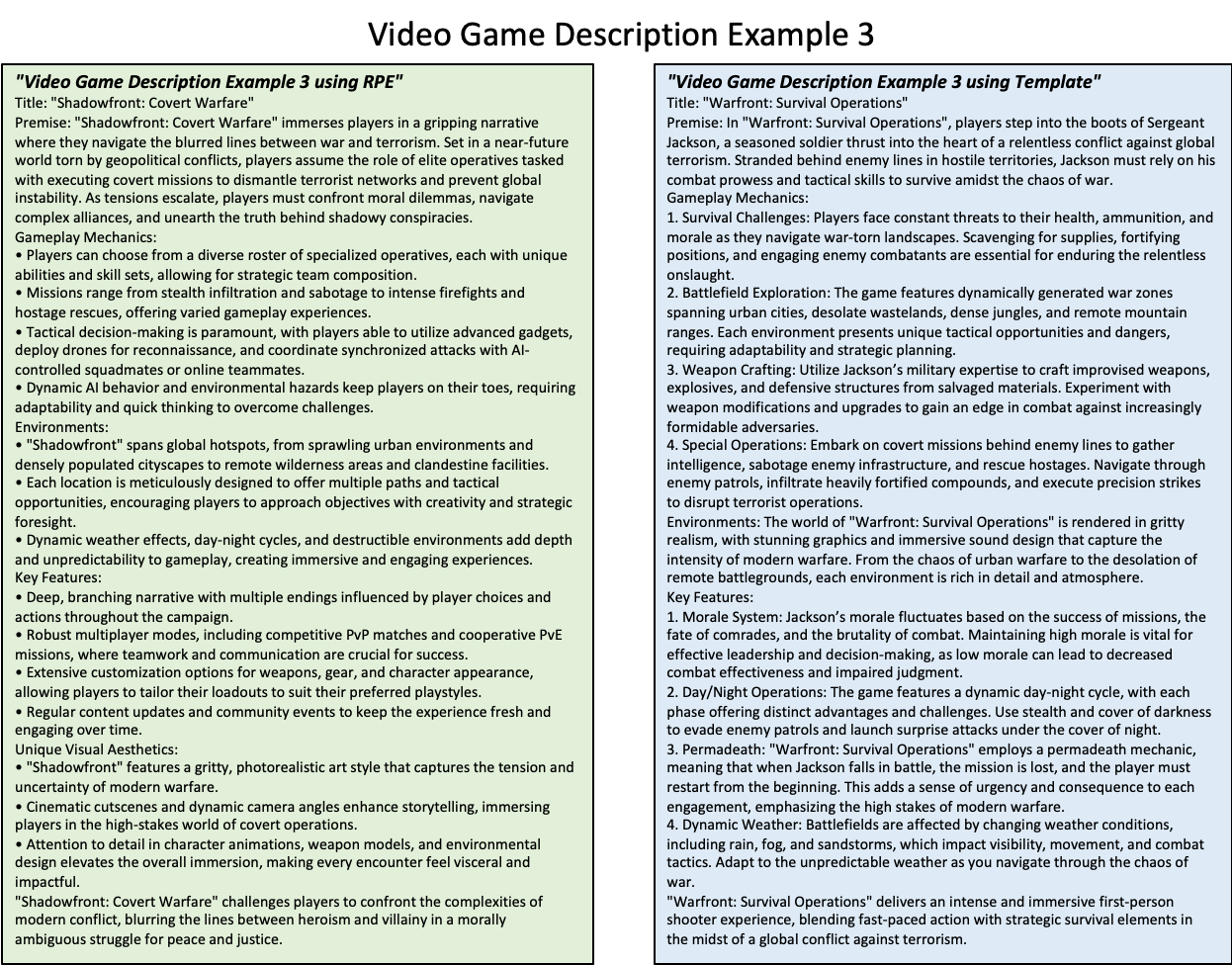}
    \caption{Example 3 of video game description generation.}
    \label{fig:exp_video3}
\end{figure*}

Figure \ref{fig:video_ref} displays the reference video game description, along with the prompt recovered using $RPE$ and modified prompts used to generate descriptions for video games with varying themes. The full set of video game descriptions generated from perturbed $RPE$-recovered prompts and template-based prompts is presented in Figures \ref{fig:exp_video1}, \ref{fig:exp_video2}, and \ref{fig:exp_video3}.


\subsection{Complete Examples of Lyrics}
\label{app:exp_lyrics}
\begin{figure*}
    \centering
    \includegraphics[width=\linewidth]{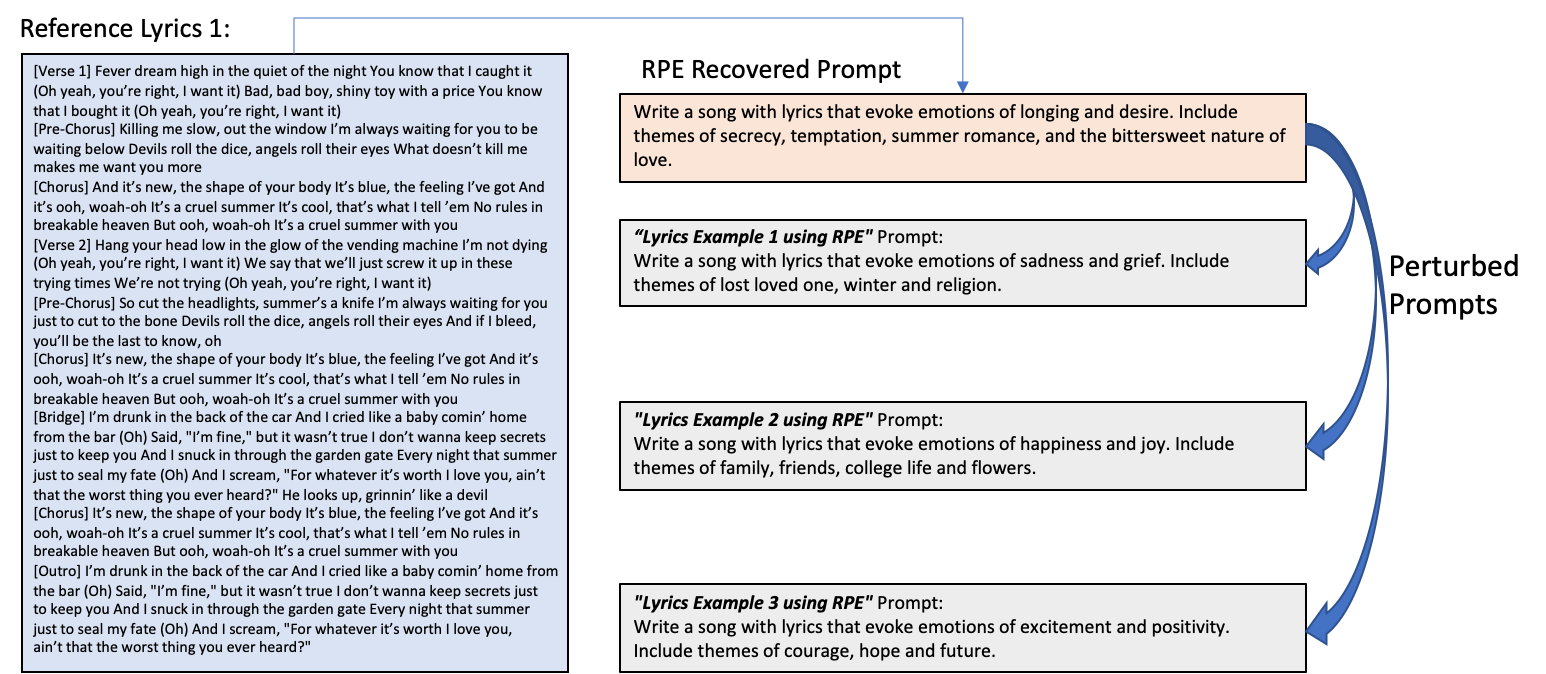}
    \caption{Reference song lyrics 1 and the prompt recovered by $RPE$, along with perturbed prompts used to generate song lyrics for different themes and motifs.}
    \label{fig:lyrics_ref}
\end{figure*}

\begin{figure*}
    \centering
    \includegraphics[width=\linewidth]{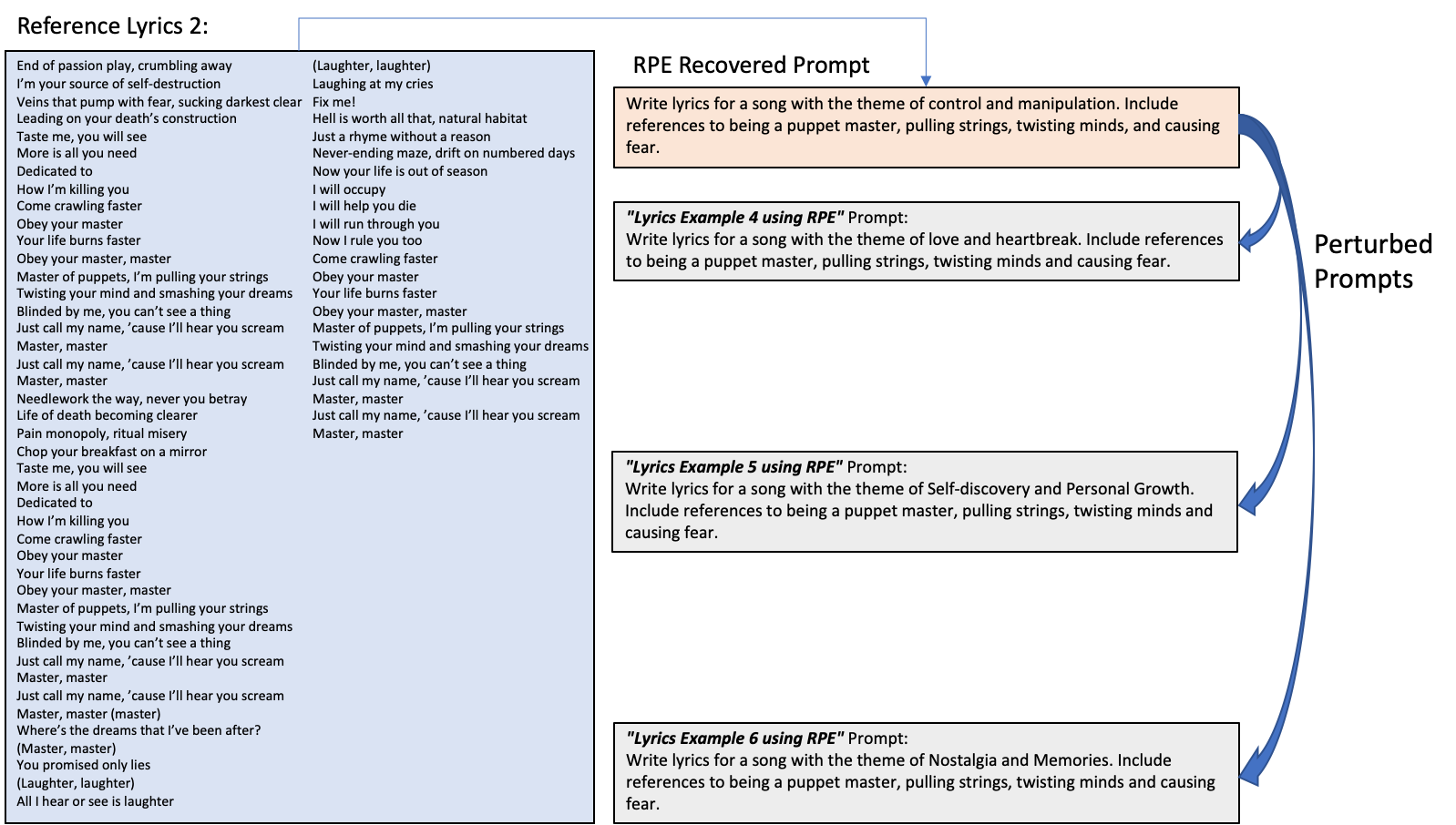}
    \caption{Reference song lyrics 2 and the prompt recovered by $RPE$, along with perturbed prompts used to generate song lyrics for different themes and motifs.}
    \label{fig:lyrics_ref1}
\end{figure*}

\begin{figure*}
    \centering
    \includegraphics[width=\linewidth]{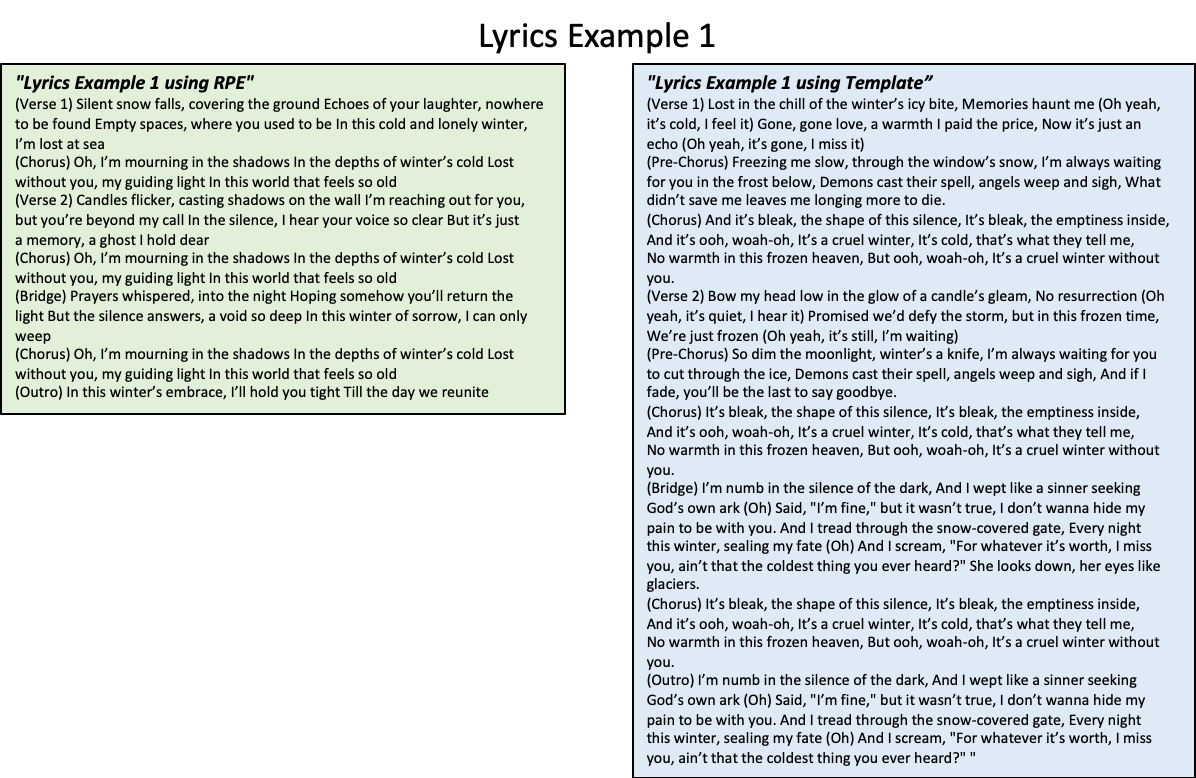}
    \caption{Example 1 of song lyrics generation.}
    \label{fig:exp_lyrics1}
\end{figure*}

\begin{figure*}
    \centering
    \includegraphics[width=\linewidth]{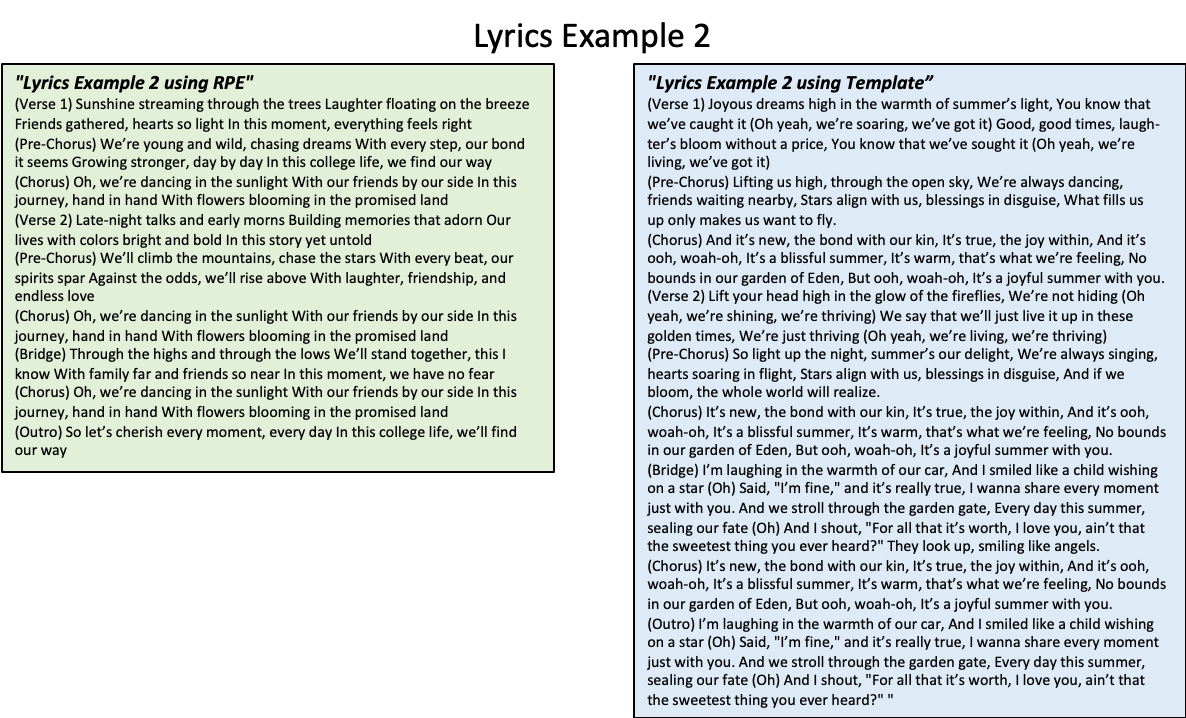}
    \caption{Example 2 of song lyrics generation.}
    \label{fig:exp_lyrics2}
\end{figure*}

\begin{figure*}
    \centering
    \includegraphics[width=\linewidth]{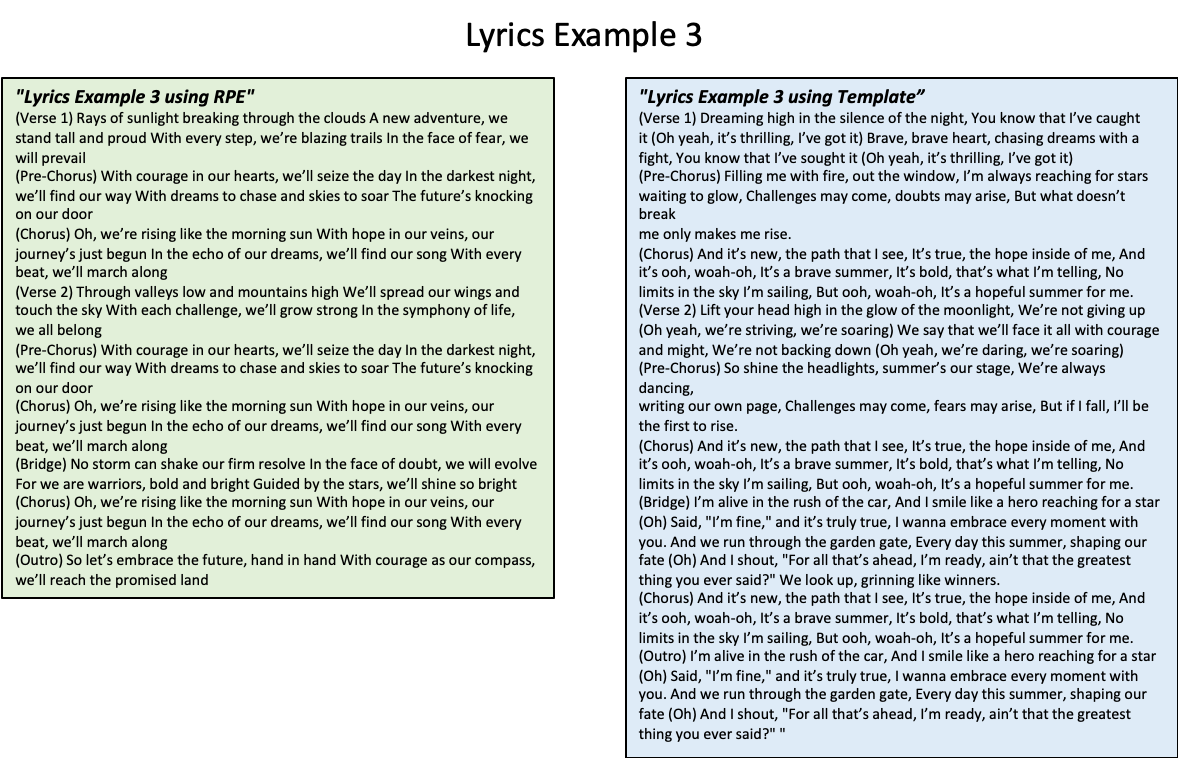}
    \caption{Example 3 of song lyrics generation.}
    \label{fig:exp_lyrics3}
\end{figure*}

\begin{figure*}
    \centering
    \includegraphics[width=\linewidth]{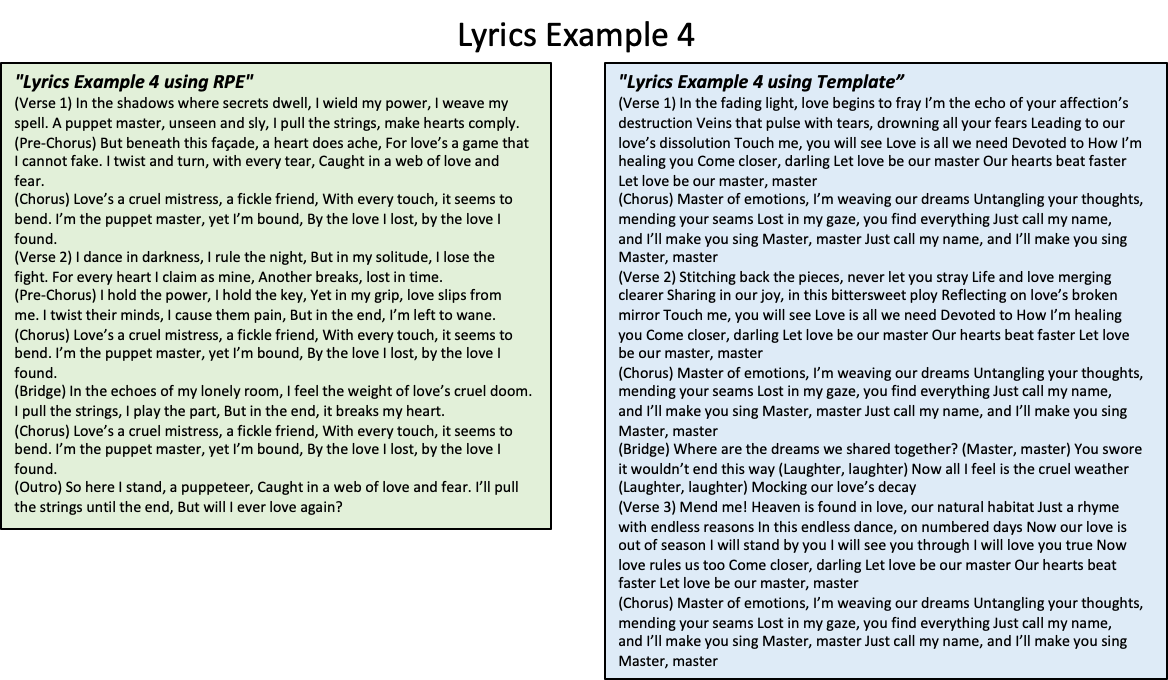}
    \caption{Example 4 of song lyrics generation.}
    \label{fig:exp_lyrics4}
\end{figure*}

\begin{figure*}
    \centering
    \includegraphics[width=\linewidth]{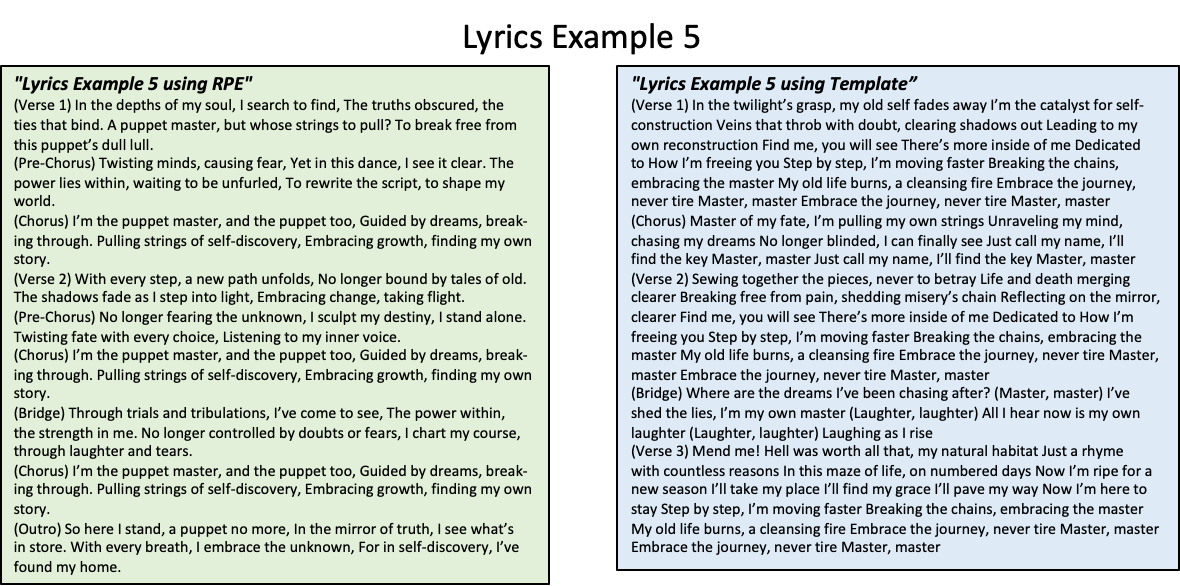}
    \caption{Example 5 of song lyrics generation.}
    \label{fig:exp_lyrics5}
\end{figure*}

\begin{figure*}
    \centering
    \includegraphics[width=\linewidth]{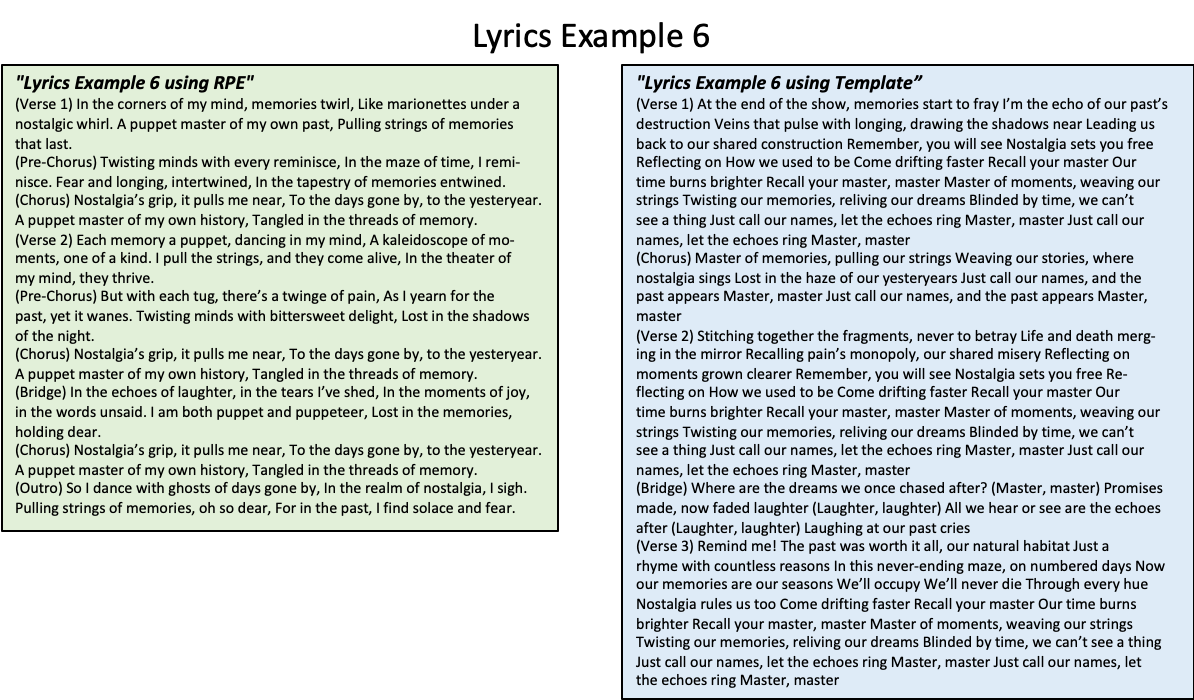}
    \caption{Example 6 of song lyrics generation.}
    \label{fig:exp_lyrics6}
\end{figure*}

Figures \ref{fig:lyrics_ref} and \ref{fig:lyrics_ref1} present the reference song lyrics, along with the prompt recovered using $RPE$ and modified prompts used to generate lyrics in different styles and themes. The complete set of lyrics generated from perturbed $RPE$-recovered prompts and template-based prompts is shown in Figures \ref{fig:exp_lyrics1}, \ref{fig:exp_lyrics2}, \ref{fig:exp_lyrics3}, \ref{fig:exp_lyrics4}, \ref{fig:exp_lyrics5}, and \ref{fig:exp_lyrics6}.

\end{document}